\documentclass[runningheads]{llncs}
\usepackage{graphicx}
\usepackage{array}
\usepackage{booktabs}

\usepackage[figuresright]{rotating}
\usepackage{xcolor}
\usepackage{float}
\usepackage{longtable}
\usepackage{setspace}
\usepackage{hyperref}
\usepackage{tikz}

\begin{document}
\title{On the Value of Labeled Data and Symbolic Methods for Hidden Neuron Activation Analysis}
\titlerunning{Symbolic Methods for Hidden Neuron Activation Analysis}
%
\author{Abhilekha Dalal\inst{1} \and
Rushrukh Rayan\inst{1} \and
Adrita Barua\inst{1} \and
Eugene Y. Vasserman\inst{1} \and
Md Kamruzzaman Sarker\inst{2} \and
Pascal Hitzler\inst{1}}
\authorrunning{A. Dalal et al.}
%
\institute{Kansas State University, Manhattan, KS, USA \and Bowie State University, Prince George's County, MD, USA}
\maketitle              
\setlength{\tabcolsep}{4pt}
\begin{abstract}

A major challenge in Explainable AI is in correctly interpreting activations of hidden neurons: accurate interpretations would help answer the question of what a deep learning system internally detects as relevant in the input, demystifying the otherwise black-box nature of deep learning systems. The state of the art indicates that hidden node activations can, in some cases, be interpretable in a way that makes sense to humans, but systematic automated methods that would be able to hypothesize and verify interpretations of hidden neuron activations are underexplored. This is particularly the case for approaches that can both draw explanations from substantial background knowledge, and that are based on inherently explainable (symbolic) methods.

In this paper, we introduce a novel model-agnostic post-hoc Explainable AI method demonstrating that it provides meaningful interpretations. Our approach is based on using a Wikipedia-derived concept hierarchy with approximately 2 million classes as background knowledge, and utilizes OWL-reasoning-based Concept Induction for explanation generation. Additionally, we explore and compare the capabilities of off-the-shelf pre-trained multimodal-based explainable methods.

Our results indicate that our approach can automatically attach meaningful class expressions as explanations to individual neurons in the dense layer of a Convolutional Neural Network. Evaluation through statistical analysis and degree of concept activation in the hidden layer show that our method provides a competitive edge in both quantitative and qualitative aspects compared to prior work.


\end{abstract}



\section{Introduction}


The explainability of the decision-making process within Deep Learning solutions is a crucial element for their integration into safety-critical systems. Robust Explainable AI techniques not only provide insights into the operations of hidden layer neurons, which significantly impact model outcomes, but also facilitate network debugging, causal factor analysis, and adjustments to data and network architecture, among other areas, to enhance model performance.

Standard assessments of deep learning performance consist of statistical evaluation, but do not seem sufficient as they cannot provide reasons or explanations for particular system behaviors~\cite{doran2018does}. While there has been significant progress in the Explainable AI domain (see Section~\ref{sec:literature_review}), the current state of the art is mostly restricted to explanation analyses based on a relatively small number of predefined explanation categories. This is problematic from a principled perspective, as it relies on the assumption that explanation categories pre-selected by humans would be viable explanation categories for deep learning systems -- an as-yet unfounded conjecture. Other approaches rely on deep learning itself, e.g., Large Language Models, to produce explanations~\cite{oikarinen2022clip} -- which means that the explanation generation method in turn is yet another black box. Other state of the art explanation systems rely on modified deep learning architectures, usually leading to a decrease in system performance compared to unmodified systems~\cite{DBLP:conf/nips/ZarlengaBCMGDSP22}. The ideal approach would exhibit strong explanation capabilities, would be self-explainable, and can take a very large pool of possible explanation categories into consideration.

Many Explainable AI techniques rely on intricate low-level data features projected into a higher-dimensional space in their explanations, limiting their accessibility to users with domain expertise \cite{lime_rebeiro,shap_lundberg,gradcam_batra}. 
Some of these methods have shown vulnerability to adversarial tampering, 
wherein altering attributed features does not prompt a change in the model's decision~\cite{attacklime_alvarez,attacklimeandshap_lakkaraju,gradcamattack_shrikumar}. Conversely, there are high-level concept-based Explainable AI approaches, where manually selected concepts are measured for their correlation with model outcomes. However, a significant question remains unanswered: whether the limited set of chosen concepts can offer a comprehensive understanding of the model's decision-making process. 
The absence of a systematic approach to consider a wide range of potential concepts that may influence the model appears to be the bottleneck. In some techniques~\cite{oikarinen2022clip}, a list of frequently occurring English words has been utilized to represent a broad concept pool, which may suffice for general applications but lacks granularity for specialized fields like gene studies or medical diagnoses, 
as the curation of the concept pool does not provide low-level control over defining natural relationships among concepts.


Our approach is motivated by several key principles: firstly, explanations should be understandable to end-users without requiring intimate familiarity with deep learning models. 
Secondly, there should be a systematic organization of human-understandable concepts with well-defined relationships among them. The extraction of relevant concepts for explaining a deep learning model's decision-making process from this defined concept pool should be automatic, thus eliminating the bottleneck of manual curation prone to confirmation bias. Another significant goal is that the explanation generation technique itself should be inherently interpretable, avoiding the use of black-box methods.
Concretely, we address herein the following common shortcomings in the state of the art.
\begin{enumerate}
\item[i.] Concepts should not be hand-picked in light of completeness.
\item[ii.] Concept extraction methods should be inherently explainable.
\item[iii.] Concepts should be understandable to the end-users without requiring deep learning expertise.
\item[iv.] Label Hypothesis concept pool should include meaningful relationships between concepts, that are made use of by the explanation approach. 
\end{enumerate}
We address these points by using \emph{Concept Induction} as core mechanism, which is based on formal logic reasoning (in the Web Ontology Language OWL) and has originally been developed for Semantic Web applications~\cite{DBLP:journals/ml/LehmannH10}. The benefits of our approach are: (a) it can be used on unmodified and pre-trained deep learning architectures, (b) it assigns explanation categories (i.e., class labels expressed in OWL) to hidden neurons such that images related to these labels activate the corresponding neuron with high probability, (c) it is inherently self-explanatory as it is based on deductive reasoning, and (d) it can construct labels from a very large pool of interconnected categories.

We hypothesize that a background knowledge with the skeleton of an ontology coupled with the inherently explainable deductive reasoning (Concept Induction) should be capable of generating meaningful explanations for the deep learning model we wish to explain.

To show that our approach can indeed provide meaningful explanations for hidden neuron activation, we instantiate it with a Convolutional Neural Network (CNN) architecture for image scene classification (trained on the ADE20K dataset~\cite{zhou2019semantic}) and a class hierarchy (i.e., a simple ontology) of approx.{} $2 \cdot 10^6$ classes derived from Wikipedia as the pool of explanation categories~\cite{DBLP:conf/kgswc/SarkerSHZNMJRA20}.\footnote{Source code, input data, raw result files, and parameter settings for replication are available online at \url{https://anonymous.4open.science/r/xai-using-wikidataAndEcii-91D9/}}
Our findings suggest that our method performs competitively, as assessed through Concept Activation analysis, which measures the relevance of concepts within the hidden layer activation space, and through statistical evaluation. When compared to other techniques such as CLIP-Dissect~\cite{oikarinen2022clip}, a pre-trained multimodal Explainable AI model, and GPT-4~\cite{achiam2023gpt}, an off-the-shelf Large Language Model, our approach demonstrates both strong quantitative and qualitative performance. 

Core contributions of the paper are as follows.
\begin{enumerate}
    \item A novel zero-shot model-agnostic Explainable AI method that explains existing pre-trained deep learning models through high-level human understandable concepts, utilizing symbolic reasoning over an ontology (or Knowledge Graph schema) as the source of explanation, which achieves state-of-the-art performance and is explainable by its nature.
    \item A method to automatically extract \textit{relevant concepts} through Concept Induction for any concept-based Explainable AI method, eliminating the need for manual selection of Label Hypothesis concepts.
    \item An in-depth comparison of explanation sources using statistical analysis for the hidden neuron perspective and Concept Activation analysis for the hidden layer perspective of our approach, a pre-trained multimodal Explainable AI method (CLIP-Dissect~\cite{oikarinen2022clip}), and a Large Language Model (GPT-4~\cite{achiam2023gpt}).
\end{enumerate}


The rest of the paper is organized as follows: in Section~\ref{sec:literature_review}, the recent approaches in Explainable AI domain are highlighted. In Section~\ref{sec:Method}, the detailed overview of the our approach along with the protocol followed to adapt CLIP-Dissect and GPT-4 for the use-case is presented. In addition to that, the description of the statistical evaluation protocol and Concept Activation analysis are presented. In Section~\ref{sec:results}, we highlight the key findings of our method and the comparative analysis with other approaches. Section~\ref{sec:discussion} presents the key observations and trade-offs of our approach, CLIP-Dissect, and GPT-4. Finally, Section~\ref{sec:conclusion} outlines the drawbacks and future scope of our study.
\section{Related Work}
\label{sec:literature_review}

Explaining
(interpreting, understanding, justifying) automated AI decisions has been explored from the early 1970s. With the recent advances in deep learning~\cite{lecun2015deep}, its wide usage in nearly every field, and its opaque nature make explainable AI more important than ever,
and there are multiple ongoing efforts to demystify deep learning~\cite{gunning2019xai,adadi2018peeking,minh2022explainable}. 
Existing explainable methods can be categorized based on input data (feature) understanding, e.g., feature summarizing~\cite{selvaraju2016grad,lime_rebeiro}, or based on the model's internal unit representation, e.g., node summarizing~\cite{zhou2018interpreting,bau2020understanding}. Those methods can be further categorized as model-specific~\cite{selvaraju2016grad} or model-agnostic~\cite{lime_rebeiro}. 
Another kind of approach relies on human interpretation of explanatory data returned, such as counterfactual questions~\cite{DBLP:journals/corr/abs-1711-00399}.

Model-agnostic feature attribution techniques, exemplified by LIME~\cite{lime_rebeiro} and SHAP~\cite{shap_lundberg}, aim to explain model predictions by measuring the influence of individual features. However, these methods face challenges, including instability in explanations~\cite{attacklime_alvarez} and susceptibility to biased classifiers~\cite{attacklimeandshap_lakkaraju}. 
Pixel attribution is a method tailored for image-based models to understand model predictions by attributing importance to individual pixels ~\cite{saliencymap_simonyan,gradcam_batra,smoothgrad_smilkov}. However, this approach faces notable limitations, especially with ReLU activation when activation becomes capped at zero~\cite{gradcamattack_shrikumar} and adversarial perturbations~\cite{unreliablepixelattribution_kim}, as even slight alterations to an image can result in varying pixel attributions, introducing instability and inconsistency in the interpretability.


\cite{tcav_kim,tcar_crabbe} came up with explanations from hand-picked concepts by using supervised learning. These methods use linear and non-linear classifiers on target concepts and use weights as CAVs. While these requires concepts to be picked manually,~\cite{ace_ghorbani} uses image segmentation and clustering to curate concepts. However, due to segmentation and clustering, some information may get lost. Another downside of the method is that -- it can only work with concepts that are visibly present in the images.~\cite{invertibletcav_zhang} proposed improvements that address the information loss and provide a way to transform explanation back to the target model's prediction by using Non-negative Matrix Factorization.

Individual Conditional Expectation (ICE) plots~\cite{ice_goldstein} and Partial Dependency Plot~\cite{pdp_friedman} offer local and global perspectives, respectively, on how predictions relate to individual features. Nevertheless, they fall short of capturing complex relationships between multiple features.

We focus on the understanding of internal units of the neural network-based deep learning models. Given a deep learning model and its prediction, we ask the questions ``What does the deep learning model's internal unit represent? Are those units activated by human-understandable concepts?''
Prior work has shown that internal units may indeed represent human-understandable concepts~\cite{zhou2018interpreting,bau2020understanding}, but these approaches require semantic segmentation~\cite{xiao2018unified}  (which is time- and compute-expensive) or explicit concept annotations~\cite{tcav_kim} (which are expensive to acquire).
To bypass these limitations, we use a statistical evaluation and concept activation analysis approach based on Concept Induction analysis for hypothesis generation (details in Section~\ref{sec:Method}). The use of large-scale
OWL background knowledge means that we draw explanations from a very large pool of explanation categories.

There has been some work using Semantic Web data to produce explanations for deep learning models~\cite{confalonieri2021using,diaz2022explainable}, and also on using Concept Induction to provide explanations~\cite{sarker2017explaining,9736291}, but they focused on analysis of input-output behavior, i.e., on generating an explanation for the overall system. We focus instead on the different task of understanding internal (hidden) node activations.

The work most similar to ours in spirit, demonstrated on a similar use case, but using a completely different approach, comes from CLIP-Dissect~\cite{oikarinen2022clip} who use the CLIP pre-trained model, employing zero-shot learning to associate images with labels. Label-Free Concept Bottleneck Models~\cite{oikarinen2023label} builds upon CLIP-Dissect and uses GPT-4~\cite{achiam2023gpt} for concept set generation.
CLIP-Dissect, as presented, has limitations that may be non-trivial to overcome without significant changes to the approach,
such as limited accuracy when predicting output labels based on concepts in the last hidden layer, not outputting a low-level concept when required (e.g., Zebra when seeing a Stripe), difficulty in transferring to other modalities or domain-specific applications (lacking an off-the-shelf concept set and the domain-specific pre-trained model). The Label-Free
approach inherits the limitations of CLIP-Dissect, and appears to be somewhat self-defeating in terms of explainability, as it uses a concept derivation method which itself is \emph{not} explainable.

In our study, we have explored the potential of Concept Induction in conjunction with large-scale background knowledge for explanation generation. This approach, aimed at understanding internal node activations, appears to be a unique contribution to the field. While other methods such as those mentioned in~\cite{oikarinen2022clip,oikarinen2023label}
have their merits, Concept Induction offers a different perspective by providing background concepts and OWL (symbolic) reasoning. It also offers a range of concept outputs, catering to different abstraction levels of concepts, and is not strictly reliant on a pre-existing, relatively small concept set or a domain-specific pre-trained concept generation model.

We have also experimented with generating relevant concepts using models trained on text and images like CLIP~\cite{clip}, and leveraging large language models like GPT-4~\cite{achiam2023gpt}. Our focus has been on quantifying the prominence of the concepts obtained from Concept Induction, CLIP, and GPT-4 within the hidden layer activation space. This has been achieved through statistical evaluation and concept activation analysis.

Our method, encompassing training, Concept Induction analysis, and evaluation analysis, is designed to be flexible and fully automatable, with the exception of the selection and provision of a suitable choice of ontology as background knowledge. It has the potential to be applied to either pre-trained models or models trained from scratch and could possibly be adapted to any neural network architecture.

\section{Method}
\label{sec:Method}

\begin{figure}[tb]
\includegraphics[width=\textwidth]{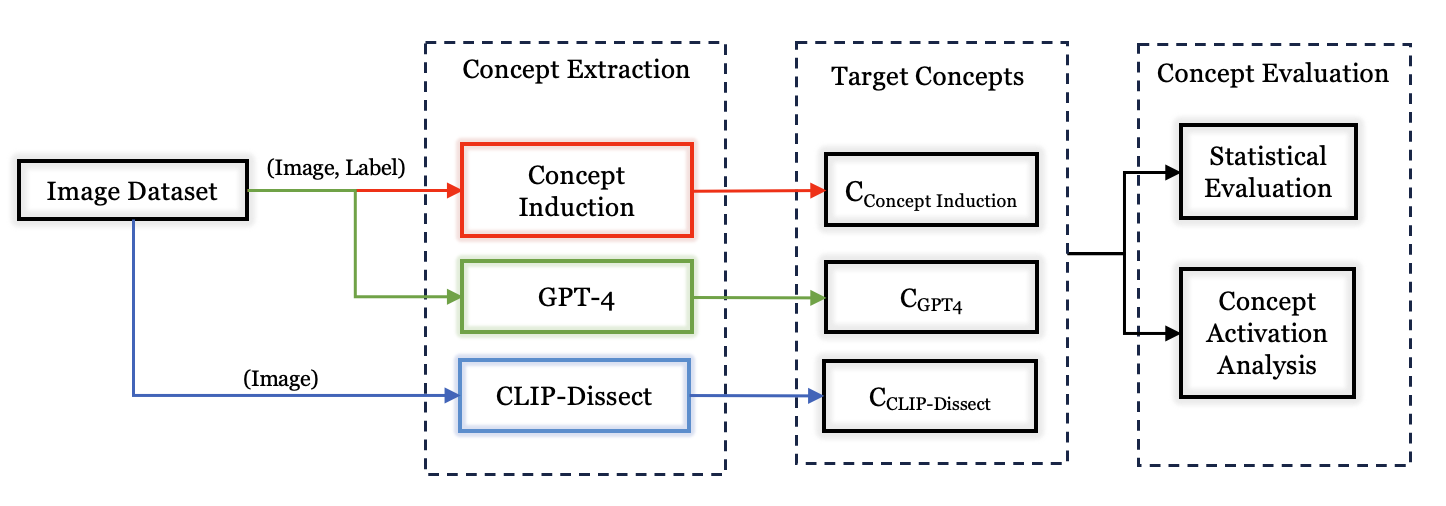}
\caption{An overview of the complete pipeline explored in this paper where Concept Extraction outlines the methods used to extract Target Concepts and Concept Evaluation outlines the evaluation methods.} \label{fig:workflow}
\end{figure}

As part of the entire work, we explore and evaluate 3 concrete methods to generate high-level concepts such that those concepts potentially provide insights into hidden layer activation space. Fig. \ref{fig:workflow} is a high-level depiction of our workflow. Components are further discussed below and throughout the paper.

\subsubsection{Preparations: Scenario and CNN Training}
\label{subsubsec:training}
We use a scene classification from images scenario
to demonstrate our approach, drawing from the ADE20K dataset~\cite{zhou2019semantic} which
contains more than 27,000 images over 365 scenes, extensively annotated with pixel-level
objects and object part labels. \emph{The annotations are not used for CNN training}, but rather only for generating label hypotheses that we will describe in Section~\ref{subsubsec:label-hypotheses}.

We train a classifier for the following scene
categories: ``bathroom,'' ``bedroom,'' ``building facade,'' ``conference room,'' ``dining
room,'' ``highway,''\linebreak ``kitchen,'' ``living room,'' ``skyscraper,'' and ``street.'' We weigh our selection toward scene
categories which have the highest number of images and we deliberately include some scene
categories that should have overlapping annotated objects -- we believe this makes the
hidden node activation analysis more interesting. We did not conduct any experiments on any
other scene selections yet, i.e., \emph{we did not change our scene selections based on
any preliminary analyses}.

\begin{table}[tb]
\caption{Performance (accuracy) of different architectures on the ADE20K dataset. The system we used, based on performance, is \textbf{in bold}.}
\label{accuracy_of_networks}
\centering
\begin{tabular}{l|l|l}
Architectures  & Training acc & Validation acc \\
\hline
    Vgg16          & 80.05\%      & 46.22\%        \\
    InceptionV3    & 89.02\%      & 51.43\%         \\
    Resnet50       & 35.01\%      & 26.56\%        \\
    \textbf{Resnet50V2}    &\textbf{87.60\%}      &\textbf{86.46\%}  \\
    Resnet101      & 53.97\%      & 53.57\%        \\
    Resnet152V2    & 94.53\%      & 51.04\%        \\
\end{tabular}
\end{table}

We trained a number of CNN architectures in order to use the one with highest accuracy, namely Vgg16~\cite{simonyan2015very}, InceptionV3~\cite{szegedy2016rethinking} and different versions of Resnet -- Resnet50, Resnet50V2, Resnet101, Resnet152V2~\cite{he2016deep,he2016identity}.
Each neural network was fine-tuned with a dataset of 6,187 images (training and validation
set) of size $224 \times 224$ for 30 epochs with early stopping\footnote{monitor validation loss; patience 3; restore
weights}
to avoid overfitting.
We used Adam as our optimization algorithm, with a categorical cross-entropy loss function and a learning rate of 0.001.

We select Resnet50V2 because it achieves the highest accuracy (see Table~\ref{accuracy_of_networks}). Note that for our investigations, which focus on explainability of hidden neuron activations, achieving a very high accuracy for the scene classification task is not essential, but a reasonably high accuracy is necessary when considering models which would be
useful in practice.

\medskip

In the following, we will detail the different components depicted in Fig. \ref{fig:workflow}. We first describe how we use Concept Induction for the generation of explanatory concepts (Section \ref{sec:CI}). Then we detail how we use CLIP-Dissect (Section \ref{sec:Clip}) and GPT-4 (Section \ref{sec:gpt4}) for the same. We finally describe our two evaluation approaches (Section \ref{subsec:conceptevaluation}).

\subsection{Concept Induction}
\label{sec:CI}

\subsubsection{Preparations: Concept Induction and Background Knowledge Base}
\label{subsubsec:CI}

\emph{Concept Induction}~\cite{DBLP:journals/ml/LehmannH10}
is based on deductive reasoning over description logics, i.e., over logics relevant to ontologies, knowledge graphs, and generally the Semantic Web field~\cite{DBLP:books/crc/Hitzler2010,DBLP:journals/cacm/Hitzler21}; in particular, the W3C Standard Web Ontology Language OWL \cite{owl2-primer} is based on description logics.
Concept Induction has indeed already been shown, in other scenarios, to be capable of producing labels that are meaningful for humans inspecting the data~\cite{DBLP:journals/corr/abs-2209-13710}.
A Concept Induction system accepts three inputs: (1) a set of positive examples $P$, (2) a set of
negative examples $N$, and (3) a knowledge base (or ontology) $K$, all expressed as description
logic theories, 
and all examples $x\in P\cup N$ occur as individuals (constants) in $K$.
It
returns description logic class expressions $E$ such that $K\models E(p)$ for all $p \in P$ and $K \not\models E(q)$ for all $q \in N$. If no such class expressions exist, then  it returns approximations for $E$ together with a number of accuracy measures.

For scalability reasons, we use the heuristic Concept Induction system
ECII \cite{DBLP:conf/aaai/SarkerH19} together with a background knowledge base that consists
only of a hierarchy of approximately 2~million classes, curated from the Wikipedia
concept hierarchy and presented in~\cite{DBLP:conf/kgswc/SarkerSHZNMJRA20}. We use
\emph{coverage} as accuracy measure, defined as
$$\textrm{coverage}(E) = \frac{|Z_1 | + |Z_2|}{|P\cup N|},$$
where $Z_1=\{p\in P\mid K\models E(p)\}$ and $Z_2 = \{n\in N\mid K\not\models E(n)\}$;
$P$ is the set of all positive instances, $N$ is the set of all negative instances, and $K$ is the knowledge base provided to ECII as part of the input.

For the Concept Induction analysis, positive and negative example sets will contain images
from ADE20K, i.e., we need to include the images in the background knowledge by linking them
to the class hierarchy. For this, we use the object annotations available for the ADE20K
images, but only part of the annotations for simplicity and scalability. More precisely, we only use the information that certain objects (such as windows) occur in certain images, and we do not make use of any of the richer annotations
such as those related to segmentation.\footnote{In principle, complex annotations in the form of sets of OWL axioms could of course be used, if a Concept Induction system is used that can deal with them, such as DL-Learner~\cite{DBLP:journals/ml/LehmannH10}. However DL-Learner does not quite scale to our size of background knowledge and task~\cite{sarker2017explaining}.} All objects from all images are then mapped to classes in the class hierarchy using the Levenshtein string similarity metric~\cite{DBLP:journals/iandc/Levenshtein75} with edit distance $0$. Mapping is in fact automated using the ``combine ontologies'' function of ECII.

For example, \texttt{ADE\_train\_00001556.jpg} in the ADE20K image set has ``door'' listed as an
object, which is mapped to the ``door'' concept in the Wikipedia concept hierarchy.
Note that the scene information is not used for the mapping, i.e., the images themselves are not assigned to specific (scene) classes in the class hierarchy -- they are connected to the hierarchy only through the objects that are shown (and annotated) in each image.

\subsubsection{Generating Label Hypotheses}
\label{subsubsec:label-hypotheses}
The general idea for generating label hypotheses using Concept Induction is as follows: given a hidden neuron,
$P$ is a set of inputs (i.e., in this case, images) to the deep learning system that activate the neuron, and $N$ is a set of inputs that do not activate the neuron
(where $P$ and $N$ are the sets of positive and negative examples, respectively).
As mentioned above, inputs are annotated with classes from the background knowledge for Concept Induction, but these annotations and the background knowledge are not part of the input to the deep learning system. ECII generates a label hypothesis\footnote{In fact, it generates several, ranked, but for now we use only the highest ranked one. We come back to this point further below.} for the given neuron on inputs $P$, $N$, and the background knowledge.

We first feed 1,370 ADE20K images to our trained Resnet50V2 and retrieve the activations of the dense layer.
We chose to look at the dense layer because previous studies indicate~\cite{distill} that earlier layers of a CNN respond to low level features such as lines, stripes, textures, colors, while layers near the final layer respond to higher-level features such as face, box, road, etc. The higher-level features align better with the nature of our background knowledge. 

The dense layer consists of 64 neurons. We chose to analyze each of the neurons separately. We are aware that activation patterns involving more than one neuron may also be informative in the sense that information may be distributed among several neurons, but the analysis of such activation patterns will be part of follow-up work.

For each neuron, we calculate the maximum activation value across all images.
We then take the positive example set $P$ to consist of all images that activate the
neuron with at least 80\% of the maximum activation value, and the negative example set $N$ to
consist of all images that activate the neuron with at most 20\% of the maximum activation
value (or do not activate it at all). The highest scoring response of running ECII on these
sets, together with the background knowledge described above, is shown in
Table~\ref{tab:listofConcepts-abbrv} for each neuron, together with the coverage of the ECII
response. For each neuron, we call its corresponding label the \emph{target label}, e.g.,
neuron $0$ has target label ``building.'' Note that some target labels consist of two  concepts, e.g., ``footboard, chain'' for neuron 49 -- this occurs if the corresponding ECII response carries two class expressions joined by a logical conjunction, i.e., in this example ``footboard $\sqcap$ chain'' (as description logic expression) or $\textnormal{footboard}(x) \land \textnormal{chain}(x)$ expressed in first-order predicate logic.

\begin{figure*}[tb]
\centering
\includegraphics[clip, trim=0in 6.3in 0in 0in, width=\textwidth]{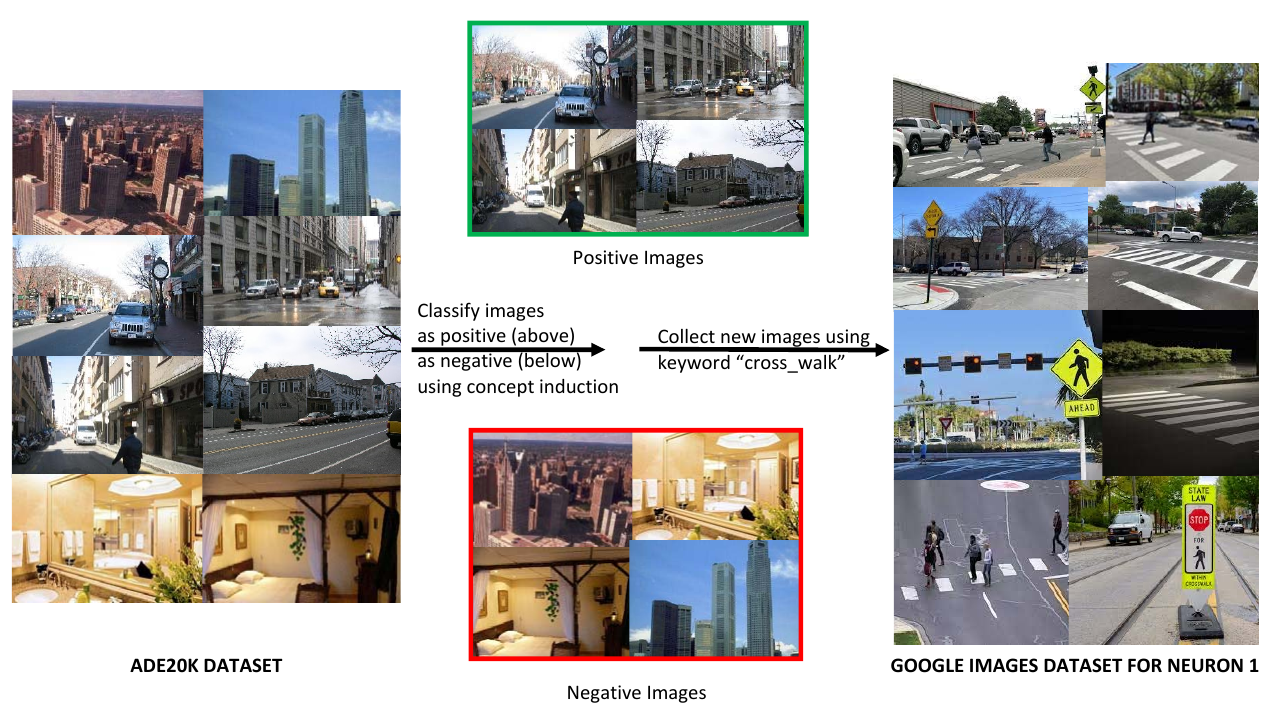} 
\caption{Example of images that were used for generating and confirming the label hypothesis for neuron 1.}
\label{fig1}
\end{figure*}

We given an example, depicted in Figure \ref{fig1}, for neuron 1. The green and red boxed images show positive and negative examples for neuron 1. Concept Induction provides "cross\_walk" as the target label. The example will be continued below.

\subsection{CLIP-Dissect}
\label{sec:Clip}

CLIP-Dissect~\cite{oikarinen2022clip} is a zero-shot Explainable AI method that associates high-level concepts with individual neurons in a designated layer. Unlike traditional training-based approaches, it utilizes the pre-trained multimodal model CLIP to gauge the resemblance between a specified set of concepts and a set of test images. Using Weighted Pointwise Mutual Information, it assesses the similarities between concepts and images in the hidden layer activation space to assign a concept to a neuron. Additionally, CLIP-Dissect is label-free, meaning that it does not require the image annotations. 

CLIP-Dissect utilizes a predefined set of the most common 20,000 English vocabulary words as concepts. To assign labels to neurons in the network under examination, we collect activations from a ResNet50v2 trained model for the ADE20K test dataset images. This results in a matrix of dimensions (Number of Images $\times$ 64), where there are 64 hidden neuron units, and each row in the matrix represents an image through its 64 hidden neuron activation values. Combining the hidden layer representations of images with the concept set as input, CLIP-Dissect assigns a label to each neuron such that the neuron is most activated when the corresponding concept is present in the image. For the ADE20K test dataset, this process generates 22 unique concepts for the 64 neurons, with several neurons having duplicate concepts.

\subsection{GPT-4}
\label{sec:gpt4}
  
We employ a Large Language Model (LLM) using a methodology similar to our previous work~\cite{barua2024concept}. Specifically, we utilize GPT-4, which represents the latest advancement in generative models and offers improved reliability, outperforming existing LLMs across various tasks~\cite{achiam2023gpt}. 
These models appear capable of generating concepts essential for distinguishing between different image classes when prompted effectively~\cite{oikarinen2023label}.

For this approach, we use the same positive ($P$) and negative ($N$) example sets from Section \ref{sec:CI}, with some minor adjustments: For Concept Induction, the negative example set ($N$) comprise all images that 
activate the neuron with at most 20\% of the maximum activation value.
Due to constraints on having a large number of negative image tags as input to GPT-4, we select only one image per class of images for each neuron to create the negative example set ($N$). The positive image set ($P$) remain unchanged, given its smaller size. All these images are sourced from the ADE20K dataset as before and are labeled with object tags present in the image.

Object tags from these images are passed into GPT-4 via the OpenAI API using prompts to generate explanations aimed at discerning the distinguishing features present in the positive set ($P$) that were absent in the negative set ($N$). These explanations were treated as concepts, and we generated a top-three list of concepts for each neuron using zero-shot prompting. For each neuron, we ran the prompt with the following parameters:

\begin{itemize}
    \item Positive example set: object tags of all positive images ($P$)
    \item Negative example set: object tags of all negative images ($N$)
    \item Prompt question: Generate the top three classes of objects or general scenario that better represent what images in the positive set ($P$) have but the images in the negative set ($N$) do not.
\end{itemize}

We employ the most recent version of the GPT-4 model for this task, with the model's temperature set to 0 and top\_p to 1. These parameters significantly influence the output diversity of GPT-4: higher temperatures (e.g., 0.7) lead to more varied and imaginative text, whereas lower temperatures (e.g., 0.2) produce more focused and deterministic responses. Setting the temperature to 0 theoretically selects the most probable token at each step, with minor variations possible due to GPU computation nuances even under deterministic settings. In contrast to temperature sampling, which modulates randomness in token selection, top\_p sampling restricts token selection to a subset (the nucleus) based on a cumulative probability mass threshold (top\_p). OpenAI's documentation advises adjusting either temperature or top\_p but not both simultaneously to control model behavior effectively. For our study, setting the temperature to 0 ensured consistency and reproducibility across outputs. More detailed information regarding the experimental setup and complete prompt can be found in~\cite{barua2024concept}. 

Although three concepts were generated for each neuron, we selected only one concept per neuron for analysis, resulting in 64 unique concepts, with several neurons having duplicate concepts.

%

\subsection{Concept Evaluation}
\label{subsec:conceptevaluation}

\subsubsection{Confirming Label Hypotheses}
\label{subsubsec:label-validation}
\label{subsubsec:label-confirmation}
The three approaches described above produce label hypotheses for all investigated neurons -- hypotheses that we will confirm or reject by testing the labels with new images. 
We use each of the target labels to search Google Images with the labels as keywords (requiring responses to be returns for \emph{both} keywords if the label is a conjunction of classes, for Concept Induction). We call each such image a \emph{target image} for the corresponding label or neuron. We use Imageye\footnote{https://chrome.google.com/webstore/detail/image-downloader-imageye/agionbommeaifngbhincahgmoflcikhm} to automatically retrieve the images, collecting up to 200 images that appear first in the Google Images search results, filtering for images in JPEG format and with a minimum size of 224x224 pixels (conforming to the size and format of ADE20K images). 

For each retrieval label, we use 80\% of the obtained images, reserving the remaining 20\% for the statistical evaluation described later in the section. The number of images used in the hypothesis confirmation step, for each label, is given in the tables. 
These images are fed to the network to check (a) whether the target neuron (with the retrieval label as target label) activates, and (b) whether any other neurons activate. The Target \% column of Tables~\ref{tab:listofConcepts-abbrv},~\ref{tab:listofConcepts-abbrvClip}, and~\ref{tab:listofConcepts-abbrvGPT} show the percentage of the target images that activate each neuron. 

Returning to our example neuron 1 in the Concept Induction case (Fig. \ref{fig1}), 88.710\% of the images retrieved with the label ``cross\_walk'' activate it.  
However, this neuron activates only for 28.923\% (indicated in the Non-Target \% column) of images retrieved using all other labels excluding ``cross\_walk.''

We define a target label for a neuron to be \emph{confirmed} if it activates 
for $\ge 80\%$ of its target images regardless of
how much or how often it activates for non-target images. 
The cut-offs for neuron activation and label hypothesis confirmation are
chosen to
ensure strong association and responsiveness to images retrieved under the target label, but 80\% is somewhat arbitrary
and could be chosen differently.

For our example neuron 1, we retrieve 233 new images with the keyword ``cross\_walk,'' 186 of which (80\%)
are used in this step. 
165 of these images, i.e., 88.710\% activate neuron 1. 
Since $88.710 \geq 80$, we consider the label
``cross\_walk'' confirmed for neuron 1.

After this step, we arrive at a list of
19 (distinct) \emph{confirmed} labels from Concept-Induction, 5 (distinct) \emph{confirmed} labels from CLIP-Dissect, and 14 (distinct) \emph{confirmed} labels from GPT-4, as listed in Tables~\ref{tab:evaluationCI},~\ref{tab:evaluationClip},~\ref{tab:evaluationGPT}.

\subsubsection{Statistical Evaluation}
\label{resultsDiscussion}
\label{subsubsec:eval}
After generating the confirmed labels (as above), we 
evaluate the node labeling using the remaining images from those retrieved from Google Images as described earlier. Results are shown in Table~\ref{tab:evaluationCI}, omitting neurons that were not activated
by any image, i.e., their maximum activation value was 0.

We consider each neuron-label pair (rows in Table~\ref{tab:evaluationCI},~\ref{tab:evaluationClip},~\ref{tab:evaluationGPT}) to be a hypothesis, e.g., for neuron 1 from Table~\ref{tab:evaluationCI}, the hypothesis is that it activates more strongly for images retrieved using the keyword ``cross\_walk'' than for images retrieved using other keywords. The corresponding null hypothesis is that activation values are \emph{not} different.
Table~\ref{tab:evaluationCI} shows the 20 hypotheses to test, corresponding to the 20 neurons with confirmed labels from method Concept Induction (recall that a double label such as neuron 16's ``mountain, bushes'' is treated as one label consisting of the conjunction of the two keywords.)

Similarly, Table~\ref{tab:evaluationClip} lists the 8 hypotheses to test, corresponding to the 8 neurons with confirmed labels from method CLIP-Dissect, and Table~\ref{tab:evaluationGPT} lists the 27 hypotheses to test, corresponding to the 27 neurons with confirmed labels from method GPT-4. 

There is no reason to assume that activation values would follow a normal distribution, or that the preconditions of the central limit theorem would be satisfied. We therefore base our statistical assessment on the Mann-Whitney U test~\cite{McKnight2010MannWhitneyUT} which is a non-parametric test that does not require a normal distribution. Essentially, by comparing the ranks of the observations in the two groups, the test allows us to determine if there is a statistically significant difference in the activation percentages between the target and non-target labels.

The resulting z-scores and p-values are shown in Table~\ref{tab:evaluationCI},~\ref{tab:evaluationClip},~\ref{tab:evaluationGPT} and are further discussed in Section~\ref{sec:results}. For our running example (neuron 1), we analyze the remaining 47 target images (20\% of the images retrieved during the label hypothesis confirmation step). Of these, 43 (91.49\%) activate the neuron with a mean and median activation of 4.17 and 4.13, respectively. Of the remaining (non-target) images in the evaluation (the sum of the image column in Table~\ref{tab:evaluationCI} minus 47), only 28.94\% activate neuron 1 for a mean of 0.67 and a median of 0.00. The Mann-Whitney U test yields a z-score of -8.92 and $p<0.00001$. The negative z-score indicates that the activation values for non-target images are indeed lower than for the target images, rejecting the null hypothesis.

It is instructive to have another look at our example neuron 1 for the Concept Induction case. The images depicted on the
left in Figure~\ref{fig1} -- target images not activating the neuron -- are mostly
computer-generated as opposed to photographic images as in the ADE20K dataset. The lower right
image does not actually show the ground at the crosswalk, but mostly sky and only indirect evidence for a crosswalk by
means of signage, which may be part of the reason why the neuron does not activate. 
The right-hand images are non-target images that activate the neuron. We
may conjecture that other road elements, prevalent in these pictures, may have triggered the
neuron. We also note that several images show bushes or plants -- 
particularly interesting because the ECII response with the third-highest coverage score is
``bushes, bush'' with a coverage score of 0.993 and 48.052\% of images retrieved using this
label actually activate the neuron (the second response for this neuron is also ``cross\_walk''). It appears that Concept Induction results should be further improvable by taking additional Concept Induction returns into consideration.

\subsubsection{Concept Activation Analysis}
\label{subsec:conceptactivationanalysis}

Concept Induction is a separate process from the neural network based processes. Leveraging the strength of the background knowledge, it outputs a list of high-level concepts. A question we can ask is: how can we know if it is possible to find existence or absence of such concepts in the hidden layer activation space?

To that extent, we employ \emph{Concept Activation}~\cite{tcav_kim,tcar_crabbe}, which is a concept-based explainable AI technique which works with a \emph{pre-defined} set of concepts. It attempts at explaining a pre-trained model by measuring the presence of \emph{concepts} in hidden-layer activations of a given image for a particular layer. For the purpose of comparative analysis, we evaluate all candidate concepts (label hypotheses), obtained from all three methods, through Concept Activation Analysis. Note that we do not restrict this analysis to only confirmed concepts, as the Concept Activation Analysis approach has not been developed with such a confirmation step as part of it. 

For each candidate concept, a set of images are collected using Imageye (exactly as described above) and a concept classifier (i.e. a Support Vector Machine) is trained. The dataset given to the concept classifier requires some pre-processing:

i. The dataset for one concept classifier consists of images that exhibit the presence of the concept under description and with images where the said concept is absent. As the concept classifier will output the existence or absence of a concept, we assign the images to have labels 0 (when concept is absent) and 1 (when concept is present). 

ii. Since we are interested in finding the concepts in the hidden layer activation space, not in the image pixel space, we need to transform the image pixel values to their activation values. To achieve that, the dataset is passed across the ResNet50V2 pre-trained model as it is the network we wish to explain. The activation values of each image in the dense layer is saved. If the dense layer consists of 64 neurons, then we end up with a matrix of dimensions (no. of images $\times$ 64). 

The transformed dataset is split into train (80\%) and test (20\%) datasets. Thereafter, a Support Vector Machine (SVM) is trained using the train split. We have used both linear (Concept Activation Vector, CAV) and non-linear (Concept Activation Region, CAR) kernel to see which decision boundary separates the presence/absence of a concept best. Once the concept classifier is trained, a test dataset is used to see to what extent the concept classifier can classify the presence/absence of concepts in the hidden layer activation space.

We use Concept Induction, CLIP-Dissect, and GPT-4 as Concept Extraction mechanism. Thereafter we use Concept Activation analysis to measure to what extent such concepts are identifiable in the hidden layer activation space. We adopt two different kernels through CAV and CAR to train an SVM and then test the classifiers on unseen image data. 
Tables~\ref{tab:concept_accuracy_long},~\ref{tab:concept_accuracy_combined},~\ref{tab:concept_accuracy_clip_dissect}, and~\ref{tab:concept_accuracy_gpt4} represent the test accuracies for the concepts extracted by Concept Induction, CLIP-Dissect, and GPT-4. Table~\ref{tab:mwuconceptactivation} represents the results of the Mann-Whitney U test performed over the test accuracies obtained from all 3 approaches. Table~\ref{tab:meanandmedianconceptactivation} shows the Mean, Median, and Standard Deviation of the test accuracies for each of the 3 approaches.
\section{Results}
\label{sec:results}

For the given test dataset split of ADE20K, we evaluate Concept Induction, CLIP-Dissect, and GPT-4 for extracting relevant candidate concepts. Subsequently, we conduct two analyses from different perspectives:

i. For each neuron of the dense layer, we identify the concepts that activate them the most (Statistical Evaluation).

ii. For each concept, we measure its degree of relevance across the entire dense layer activation space (Concept Activation Analysis).

The combination of these two perspectives—a detailed examination of how each neuron unit functions and a broader view of how the dense layer operates as a whole enables us to gain a comprehensive insight into the inner workings of hidden layer computations. Regarding statistical evaluation, we rigorously assess the significance of differences in activation percentages between target and non-target labels for each confirmed label hypothesis. We compute the z-score and p-value using the non-parametric Mann-Whitney U test. Additionally, we calculate the Mean and Median for both target and non-target labels to further characterize the results. In the Concept Activation Analysis, we evaluate the effectiveness of concepts across several dimensions. Initially, we assess each concept classifier considering both linear (CAV) and non-linear (CAR) decision boundary based on the presence and absence of each concept. To validate that the concept classifier's test accuracy is not merely coincidental, we conduct k-fold cross-validations and calculate p-values. Additionally, we compute the Mean, Median, and Standard Deviation, and perform the Mann-Whitney U test to quantify the statistical significance of the test accuracies. This comprehensive approach ensures a robust evaluation of the concepts' performance in activating the hidden layer.

Our findings suggest that Concept Induction consistently performs well in both sets of evaluations conducted -- Statistical Evaluation and Concept Activation Analysis (Section~\ref{subsec:conceptevaluation}).
From the statistical evaluation, it is evident that Concept Induction achieves better performance than that of CLIP-Dissect and GPT-4. In the Concept Activation Analysis, quantitative measures reveal that Concept Induction achieves comparable performance to CLIP-Dissect, with GPT-4 exhibiting the lowest performance. Conversely, the Concept Induction approach demonstrates several notable qualitative advantages over both CLIP-Dissect and GPT-4:
\begin{itemize}
    \item CLIP-Dissect and GPT-4 are black-box models used as a concept extraction method to explain a probing network, which in this case is a CNN model, i.e., this approach to explainability is itself not readily explainable. In contrast, Concept Induction, serving as a concept extraction method, inherently offers explainability as it operates on deductive reasoning principles.
    \item CLIP-Dissect relies on a common English vocabulary (about 20K words) as the pool of concepts, whereas Concept Induction is supported by a meticulously constructed background knowledge (in this case with about 2M concepts), affording greater control over the definition of explanations through hierarchical relationships.
    \item While GPT-4/CLIP-Dissect emulate intuitive and rapid decision-making processes, Concept Induction follows a systematic and logical decision-mak\-ing approach -- thereby rendering our approach to be explainable by nature.
\end{itemize}

The results in Tables~\ref{tab:evaluationCI},~\ref{tab:evaluationClip},~\ref{tab:evaluationGPT} show that Concept Induction analysis with large-scale background knowledge yields meaningful labels that stably explain neuron activation. Of the 20 null hypotheses from Concept Induction, 19 are rejected at $p < 0.05$, but most (all except neurons 0, 18 and 49) are rejected at much lower p-values. Only neuron 0's null hypothesis could not be rejected. With CLIP-Dissect, all 8 null hypotheses are rejected at $p < 0.05$, and with GPT-4, 25 out of 27 null hypotheses are rejected at $p < 0.05$, with exceptions for neurons 14 and 31. Excluding repeating concepts, Concept Induction yields \textbf{19} statistically validated hypotheses, CLIP-Dissect yields \textbf{5}, and GPT-4 yields \textbf{12}.

The Non-Target \% column of Table~\ref{tab:listofConcepts-abbrv} provides some insight into the results for neurons 0, 18, 49 and neurons 14, 31 from Table~\ref{tab:listofConcepts-abbrvGPT}: target and non-target values for these neurons are closer to each other. 
Likewise, differences between target and non-target values for mean activation values and median activation values in Tables~\ref{tab:evaluationCI},~\ref{tab:evaluationGPT} are smaller for these neurons. This hints at ways to improve label hypothesis generation or confirmation, and we will discuss this and other ideas for further improvement below under possible future work.

Mann-Whitney U results show that, for most neurons listed in Tables~\ref{tab:evaluationCI},~\ref{tab:evaluationClip},~\ref{tab:evaluationGPT} (with $p<0.00001$), activation values of target images are \emph{overwhelmingly} higher than that of non-target images. The negative z-scores with high absolute values informally indicate the same, as do the mean and median values. Neurons 16 and 49 of Table~\ref{tab:evaluationCI}, for which the hypotheses also hold but with $p<0.05$ and $p<0.01$, respectively, still exhibit statistically significant higher activation values for target than for non-target images, but not overwhelmingly so. This can also be informally seen from lower absolute values of the z-scores, and from smaller differences between the means and the medians.

For the Concept Activation Analysis evaluation, Concept Induction yields \textbf{69} unique concepts with Mean Test Accuracy of \textbf{0.9154} (CAV) and \textbf{0.9150} (CAR). CLIP-Dissect identifies \textbf{22} concepts with Mean Test Accuracy of \textbf{0.9160} (CAV) and \textbf{0.9259} (CAR). GPT-4 produces \textbf{21} concepts with Mean Test Accuracy of \textbf{0.8757} (CAV) and \textbf{0.8887} (CAR). Although, based solely on the numeric values of Mean Test Accuracy, CLIP-Dissect demonstrates a slightly superior performance compared to Concept Induction, and GPT-4 performs the least, we contend that the substantially higher number of concepts generated by Concept Induction allows CLIP-Dissect to achieve a marginally higher test accuracy. By considering the top 22 (equal to the number of concepts generated by CLIP-Dissect) test accuracies of concepts extracted by Concept Induction, the Mean Test Accuracy increases to \textbf{0.9599} (CAV) and \textbf{0.9584} (CAR). For statistical confirmation, we conduct a p-value test for K-fold cross validation, wherein all concepts in Concept Activation analysis achieve $p < 0.05$. Using a Mann-Whitney U test, we statistically ascertain that CLIP-Dissect outperforms GPT-4 in terms of CAR, and Concept Induction surpasses GPT on CAV (see Table~\ref{tab:mwuconceptactivation}).

This analysis leads us to the following conclusion: Among the three approaches we evaluate, Concept Induction demonstrates superior performance both in the quantity of high-quality concepts generated and in the relevance of these concepts within the hidden layer activation space. Furthermore, our approach possesses inherent explainability as it does not depend on any pre-trained black-box model to identify candidate concepts. However, there are undoubtedly trade-offs involved in selecting among the three approaches, which we elaborate on in Section~\ref{sec:discussion}.

Based on the results obtained from the Statistical Evaluation and Concept Activation analysis, our approach introduces a novel zero-shot, model-agnostic Explainable AI technique. This technique offers insights into the hidden layer activation space by utilizing high-level, human-understandable concepts. Leveraging deductive reasoning over background knowledge, our approach inherently provides explainability while also achieving competitive performance, thus confirming our initial hypothesis.

\begin{table}
\caption{Concept Induction -- The omitted neurons were not activated
by any image, i.e., their maximum activation value was 0. Images: Number of images used per label. Target \%: Percentage of target images activating the neuron above 80\% of its maximum activation. Non-Target \%: The same, but for all other images.
\textbf{Bold} denotes the 20 neurons whose labels are considered confirmed.}
\label{tab:listofConcepts-abbrv}
\centering
\resizebox{.88\columnwidth}{!}{
\begin{footnotesize}
\begin{tabular}{clrrrr}
    Neuron & Obtained Label(s) & Images & Coverage & Target \% & Non-Target \% \\
    \hline
    \textbf{0} & \textbf{building} & \textbf{164} & \textbf{0.997} & \textbf{89.024} & \textbf{72.328} \\
    \textbf{1} & \textbf{cross\_walk} & \textbf{186} & \textbf{0.994} & \textbf{88.710} & \textbf{28.923} \\
    \textbf{3} & \textbf{night\_table} & \textbf{157} & \textbf{0.987} & \textbf{90.446} & \textbf{56.714} \\
    6 & dishcloth, toaster & 106 & 0.999 & 16.038 & 39.078 \\
    7 & toothbrush, pipage & 112 & 0.991 & 75.893 & 59.436 \\
    \hline
    \textbf{8} & \textbf{shower\_stall, cistern} & \textbf{136} & \textbf{0.995} & \textbf{100.000} & \textbf{53.186} \\
    11 & river\_water & 157 & 0.995 & 31.847 & 22.309 \\
    12 & baseboard, dish\_rag & 108 & 0.993 & 75.926 & 48.248 \\
    14 & rocking\_horse, rocker & 86 & 0.985 & 54.651 & 47.816 \\
    \textbf{16} & \textbf{mountain, bushes} & \textbf{108} & \textbf{0.995} & \textbf{87.037} & \textbf{24.969} \\
    \hline
    17 & stem & 133 & 0.993 & 30.827 & 31.800 \\
    \textbf{18} & \textbf{slope} & \textbf{139} & \textbf{0.983} & \textbf{92.086} & \textbf{69.919} \\
    \textbf{19} & \textbf{wardrobe, air\_conditioning} & \textbf{110} & \textbf{0.999} & \textbf{89.091} & \textbf{65.034} \\
    20 & fire\_hydrant & 158 & 0.990 & 5.696 & 13.233 \\
    \textbf{22} & \textbf{skyscraper} & \textbf{156} & \textbf{0.992} & \textbf{99.359} & \textbf{54.893} \\
    \hline
    23 & fire\_escape & 162 & 0.996 & 61.111 & 18.311 \\
    25 & spatula, nuts & 126 & 0.999 & 2.381 & 0.883 \\
    26 & skyscraper, river & 112 & 0.995 & 77.679 & 35.489 \\
    27 & manhole, left\_arm & 85 & 0.996 & 35.294 & 26.640 \\
    28 & flooring, fluorescent\_tube & 115 & 1.000 & 38.261 & 33.198 \\
    \hline
    \textbf{29} & \textbf{lid, soap\_dispenser} & \textbf{131} & \textbf{0.998} & \textbf{99.237} & \textbf{78.571} \\
    \textbf{30} & \textbf{teapot, saucepan} & \textbf{108} & \textbf{0.998} & \textbf{81.481} & \textbf{47.984} \\
    \textbf{31} & fire\_escape & 162 & 0.961 & 77.160 & 63.147 \\
    33 & tanklid, slipper & 81 & 0.987 & 41.975 & 30.214 \\
    34 & left\_foot, mouth & 110 & 0.994 & 20.909 & 49.216 \\
    \hline
    35 & utensils\_canister, body & 111 & 0.999 & 7.207 & 11.223 \\
    \textbf{36} & \textbf{tap, crapper} & \textbf{92} & \textbf{0.997} & \textbf{89.130} & \textbf{70.606} \\
    37 & cistern, doorcase & 101 & 0.999 & 21.782 & 24.147 \\
    38 & letter\_box, go\_cart & 125 & 0.999 & 28.000 & 31.314 \\
    39 & side\_rail & 148 & 0.980 & 35.811 & 34.687 \\
    \hline
    40 & sculpture, side\_rail & 119 & 0.995 & 25.210 & 21.224 \\
    \textbf{41} & \textbf{open\_fireplace, coffee\_table} & \textbf{122} & \textbf{0.992} & \textbf{88.525} & \textbf{16.381} \\
    42 & pillar, stretcher & 117 & 0.998 & 52.137 & 42.169 \\
    \textbf{43} & \textbf{central\_reservation} & \textbf{157} & \textbf{0.986} & \textbf{95.541} & \textbf{84.973} \\
    44 & saucepan, dishrack & 120 & 0.997 & 69.167 & 36.157 \\
    \hline
    46 & Casserole & 157 & 0.999 & 45.223 & 36.394 \\
    \textbf{48} & \textbf{road} & \textbf{167} & \textbf{0.984} & \textbf{100.000} & \textbf{73.932} \\
    \textbf{49} & \textbf{footboard, chain} & \textbf{126} & \textbf{0.982} & \textbf{88.889} & \textbf{66.702} \\
    50 & night\_table & 157 & 0.972 & 65.605 & 62.735 \\
    \textbf{51} & \textbf{road, car} & \textbf{84} & \textbf{0.999} & \textbf{98.810} & \textbf{48.571} \\
    \hline
    53 & pylon, posters & 104 & 0.985 & 11.538 & 17.332 \\
    \textbf{54} & \textbf{skyscraper} & \textbf{156} & \textbf{0.987} & \textbf{98.718} & \textbf{70.432} \\
    \textbf{56} & \textbf{flusher, soap\_dish} & \textbf{212} & \textbf{0.997} & \textbf{90.094} & \textbf{63.552} \\
    \textbf{57} & \textbf{shower\_stall, screen\_door} & \textbf{133} & \textbf{0.999} & \textbf{98.496} & \textbf{31.747} \\
    58 & plank, casserole & 80 & 0.998 & 3.750 & 3.925 \\
    \hline
    59 & manhole, left\_arm & 85 & 0.994 & 35.294 & 21.589 \\
    60 & paper\_towels, jar & 87 & 0.999 & 0.000 & 1.246 \\
    61 & ornament, saucepan & 102 & 0.995 & 43.137 & 17.274 \\
    62 & sideboard & 100 & 0.991 & 21.000 & 29.734 \\
    \textbf{63} & \textbf{edifice, skyscraper} & \textbf{178} & \textbf{0.999} & \textbf{92.135} & \textbf{48.761}\\
 
    \hline
\end{tabular}
\end{footnotesize}}
\end{table}

\begin{table}
\caption{CLIP-Dissect -- The omitted neurons were not activated by any image, i.e., their maximum activation value was 0. Images: Number of images used per label. Target \%: Percentage of target images activating the neuron above 80\% of its maximum activation. Non-Target \%: The same, but for all other images.
\textbf{Bold} denotes the 8 neurons whose labels are considered confirmed.}
\label{tab:listofConcepts-abbrvClip}
\centering
\resizebox{.70\columnwidth}{!}{
\begin{footnotesize}
\begin{tabular}{clrrrr}
Neuron & Obtained Label(s) & Images & Target \% & Non-target\% \\
\hline
0 & restaurants & 140 & 55.000 & 59.295 \\
1 & restaurants & 140 & 32.143 & 33.851 \\
\textbf{3} & \textbf{dresser} & \textbf{171} & \textbf{95.322} & \textbf{66.199} \\
6 & dining & 153 & 7.190 & 50.195 \\
\textbf{7} & \textbf{bathroom} & \textbf{153} & \textbf{93.333} & \textbf{44.113} \\
\hline
8 & restaurants & 140 & 24.286 & 37.957 \\
11 & highway & 153 & 14.063 & 25.153 \\
12 & street & 140 & 5.797 & 50.253 \\
14 & file & 160 & 54.375 & 69.867 \\
16 & bathroom & 171 & 2.000 & 31.722 \\
\hline
17 & furnished & 169 & 62.130 & 36.390 \\
\textbf{18} & \textbf{dining} & \textbf{153} & \textbf{93.464} & \textbf{74.448} \\
19 & bathroom & 149 & 77.333 & 56.471 \\
20 & buildings & 107 & 13.725 & 19.610 \\
22 & road & 258 & 51.550 & 46.487 \\
23 & bedroom & 123 & 0.637 & 18.823 \\
\hline
25 & restaurants & 140 & 12.857 & 5.044 \\
26 & restaurants & 140 & 2.143 & 44.552 \\
27 & bedroom & 150 & 2.548 & 27.763 \\
28 & dining & 153 & 9.150 & 40.747 \\
29 & street & 150 & 78.261 & 66.277 \\
\hline
30 & bed & 150 & 29.375 & 36.154 \\
31 & mississauga & 146 & 30.137 & 57.175 \\
\textbf{33} & \textbf{bathroom} & \textbf{150} & \textbf{80.667} & \textbf{32.955} \\
34 & microwave & 102 & 3.922 & 50.240 \\
35 & roundtable & 72 & 16.667 & 14.932 \\
\hline
36 & municipal & 154 & 51.299 & 67.002 \\
37 & bed & 160 & 8.125 & 17.670 \\
\textbf{38} & \textbf{bathroom} & \textbf{150} & \textbf{90.667} & \textbf{32.566} \\
39 & restaurants & 140 & 26.429 & 39.961 \\
40 & dining & 153 & 5.882 & 32.143 \\
\hline
41 & bedroom & 157 & 64.968 & 34.428 \\
42 & room & 156 & 35.897 & 45.206 \\
\textbf{43} & \textbf{highways} & \textbf{128} & \textbf{100.000} & \textbf{61.900} \\
44 & buildings & 153 & 9.150 & 38.377 \\
46 & restaurants & 140 & 23.571 & 33.269 \\
\hline
48 & bedroom & 157 & 8.917 & 60.241 \\
\textbf{49} & \textbf{bedroom} & \textbf{157} & \textbf{95.541} & \textbf{55.917} \\
\textbf{50} & \textbf{bedroom} & \textbf{157} & \textbf{100.000} & \textbf{62.744} \\
51 & bedroom & 157 & 4.459 & 51.951 \\
53 & kitchens & 155 & 50.968 & 24.886 \\
\hline
54 & dining & 153 & 13.725 & 62.857 \\
56 & bedroom & 157 & 1.911 & 45.676 \\
58 & buildings & 153 & 0.654 & 10.455 \\
59 & buildings & 153 & 35.294 & 24.156 \\
61 & street & 69 & 1.449 & 14.697 \\
62 & street & 69 & 24.638 & 44.722 \\
63 & bathroom & 150 & 16.667 & 47.584 \\
\hline
\end{tabular}
\end{footnotesize}}
\end{table}

\begin{table}
\caption{GPT-4 -- The omitted neurons were not activated by any image, i.e., their maximum activation value was 0. Images: Number of images used per label. Target \%: Percentage of target images activating the neuron above 80\% of its maximum activation. Non-Target \%: The same, but for all other images.
\textbf{Bold} denotes the 27 neurons whose labels are considered confirmed.}
\label{tab:listofConcepts-abbrvGPT}
\centering
\resizebox{.70\columnwidth}{!}{
\begin{footnotesize}
\begin{tabular}{clrrrr}
Neuron & Obtained Label(s) & Images & Target \% & Non-target\% \\
\hline
0 & Urban Landscape & 176 & 54.545 & 59.078 \\
\textbf{1} & \textbf{Street Scene} & \textbf{164} & \textbf{92.073} & \textbf{29.884} \\
\textbf{3} & \textbf{Bedroom} & \textbf{165} & \textbf{97.576} & \textbf{62.967} \\
\textbf{6} & \textbf{Kitchen} & \textbf{171} & \textbf{86.550} & \textbf{51.733} \\
7 & Indoor Home Decor & 177 & 66.102 & 44.793 \\
\hline
\textbf{8} & \textbf{Bathroom} & \textbf{164} & \textbf{98.780} & \textbf{47.897} \\
11 & Kitchen Scene & 167 & 41.916 & 26.281 \\
12 & Indoor Home Setting & 164 & 62.805 & 47.205 \\
\textbf{14} & \textbf{Living Room} & \textbf{164} & \textbf{82.317} & \textbf{65.053} \\
16 & Urban Landscape & 176 & 73.864 & 28.290 \\
\hline
\textbf{17} & \textbf{Dining Room} & \textbf{159} & \textbf{93.711} & \textbf{46.339} \\
\textbf{18} & \textbf{Outdoor Scenery} & \textbf{164} & \textbf{92.073} & \textbf{73.852} \\
19 & Indoor Home Decor & 177 & 29.379 & 45.571 \\
20 & Street Scene & 164 & 68.902 & 14.305 \\
\textbf{22} & \textbf{Street Scene} & \textbf{164} & \textbf{90.244} & \textbf{51.273} \\
\hline
\textbf{23} & \textbf{Street Scene} & \textbf{164} & \textbf{81.098} & \textbf{19.507} \\
25 & Kitchen & 171 & 21.637 & 5.628 \\
26 & Cityscape & 156 & 73.718 & 28.023 \\
27 & Urban Transportation & 163 & 66.871 & 30.152 \\
28 & Classroom & 162 & 60.494 & 60.494 \\
\hline
\textbf{29} & \textbf{Bathroom} & \textbf{164} & \textbf{91.463} & \textbf{68.926} \\
\textbf{30} & \textbf{Kitchen} & \textbf{171} & \textbf{90.643} & \textbf{41.724} \\
\textbf{31} & \textbf{Urban Street Scene} & \textbf{163} & \textbf{80.864} & \textbf{67.201} \\
33 & Bathroom & 164 & 74.390 & 37.272 \\
34 & Eyeglasses & 168 & 65.476 & 45.208 \\
\hline
35 & Kitchen & 171 & 66.667 & 13.224 \\
\textbf{36} & \textbf{Bathroom} & \textbf{164} & \textbf{95.122} & \textbf{61.704} \\
37 & Bathroom & 164 & 43.902 & 10.487 \\
\textbf{38} & \textbf{Living Room} & \textbf{164} & \textbf{94.512} & \textbf{56.087} \\
\textbf{39} & \textbf{Bicycle} & \textbf{156} & \textbf{82.692} & \textbf{46.328} \\
\hline
40 & Living Room & 164 & 70.122 & 24.156 \\
\textbf{41} & \textbf{Living Room} & \textbf{164} & \textbf{95.122} & \textbf{41.616} \\
42 & Living Room & 164 & 48.780 & 46.431 \\
\textbf{43} & \textbf{Outdoor Urban Scene} & \textbf{163} & \textbf{91.411} & \textbf{57.925} \\
\textbf{44} & \textbf{Kitchen Scene} & \textbf{167} & \textbf{86.826} & \textbf{45.721} \\
\hline
46 & Kitchen Scene & 167 & 43.114 & 31.155 \\
\textbf{48} & \textbf{Urban Street Scene} & \textbf{163} & \textbf{99.383} & \textbf{55.061} \\
\textbf{49} & \textbf{Bedroom} & \textbf{165} & \textbf{95.758} & \textbf{36.120} \\
\textbf{50} & \textbf{Living Room} & \textbf{164} & \textbf{93.902} & \textbf{62.756} \\
\textbf{51} & \textbf{Street Scene} & \textbf{164} & \textbf{98.171} & \textbf{43.830} \\
\hline
53 & Street Scene & 164 & 57.317 & 23.575 \\
54 & Home Interior & 165 & 26.061 & 63.216 \\
\textbf{56} & \textbf{Toilet Brush} & \textbf{165} & \textbf{94.545} & \textbf{35.095} \\
\textbf{57} & \textbf{Bathroom Interior} & \textbf{165} & \textbf{95.092} & \textbf{41.549} \\
58 & Kitchen Scenario & 165 & 29.268 & 11.096 \\
\hline
\textbf{59} & \textbf{Urban Street Scene} & \textbf{163} & \textbf{87.037} & \textbf{26.217} \\
60 & Kitchen & 171 & 0.585 & 1.691 \\
61 & Kitchen & 171 & 60.819 & 11.810 \\
\textbf{62} & \textbf{Dining Room} & \textbf{159} & \textbf{94.969} & \textbf{44.128} \\
\textbf{63} & \textbf{Cityscape} & \textbf{156} & \textbf{95.513} & \textbf{47.791}\\
\hline
\end{tabular}
\end{footnotesize}}
\end{table}

\begin{table}
\caption{Concept Induction -- Evaluation details as discussed in Section~\ref{subsubsec:eval}.
Images: Number of images used for evaluation.
\# Activations: (targ(et)): Percentage of target images activating the neuron (i.e.,
activation at least 80\% of this neuron's activation maximum);
(\mbox{non-t}):
Same for all
other images used in the evaluation. Mean/Median (targ(et)/non-t(arget)): Mean/median activation value for
target and non-target images, respectively.}
\label{tab:evaluationCI}
\centering
\resizebox{\columnwidth}{!}{%
\begin{tabular}{clc|rr|rr|rr|rr}
     \multicolumn{1}{l}{Neuron} & \multicolumn{1}{l}{Label(s)} & \multicolumn{1}{c}{Images} & \multicolumn{2}{c}{\# Activations (\%)} & \multicolumn{2}{c}{Mean} & \multicolumn{2}{c}{Median} & \multicolumn{1}{c}{z-score} & \multicolumn{1}{c}{p-value}\\
     \hline
     &  & & targ & non-t & targ & non-t & targ & non-t &  & \\
    \hline
    0 & building & 42 & 80.95 & 73.40 & 2.08 & 1.81 & 2.00 & 1.50 & -1.28 & 0.0995 \\
    1 & cross\_walk & 47 & 91.49 & 28.94 & 4.17 & 0.67 & 4.13 & 0.00 & -8.92 & \textless .00001 \\
    3 & night\_table & 40 & 100.00 & 55.71 & 2.52 & 1.05 & 2.50 & 0.35 & -6.84 & \textless .00001 \\
    8 & shower\_stall, cistern & 35 & 100.00 & 54.40 & 5.26 & 1.35 & 5.34 & 0.32 & -8.30 & \textless .00001 \\
    16 & mountain, bushes & 27 & 100.00 & 25.42 & 2.33 & 0.67 & 2.17 & 0.00 & -6.72 & \textless .00001 \\
    \hline
    18 & slope & 35 & 91.43 & 68.85 & 1.59 & 1.37 & 1.44 & 1.00 & -2.03 & 0.0209 \\
    19 & wardrobe, air\_conditioning & 28 & 89.29 & 65.81 & 2.30 & 1.28 & 2.30 & 0.84 & -4.00 & \textless .00001 \\
    22 & skyscraper & 39 & 97.44 & 56.16 & 3.97 & 1.28 & 4.42 & 0.33 & -7.74 & \textless .00001 \\
    29 & lid, soap\_dispenser & 33 & 100.00 & 80.47 & 4.38 & 2.14 & 4.15 & 1.74 & -5.92 & \textless .00001 \\
    30 & teapot, saucepan & 27 & 85.19 & 49.93 & 2.52 & 1.05 & 2.23 & 0.00 & -4.28 & \textless .00001 \\
    \hline
    36 & tap, crapper & 23 & 91.30 & 70.78 & 3.24 & 1.75 & 2.82 & 1.29 & -3.59 & \textless .00001 \\
    41 & open\_fireplace, coffee\_table & 31 & 80.65 & 15.11 & 2.03 & 0.14 & 2.12 & 0.00 & -7.15 & \textless .00001 \\
    43 & central\_reservation & 40 & 97.50 & 85.42 & 7.43 & 3.71 & 8.08 & 3.60 & -5.94 & \textless .00001 \\
    48 & road & 42 & 100.00 & 74.46 & 6.15 & 2.68 & 6.65 & 2.30 & -7.78 & \textless .00001 \\
    49 & footboard, chain & 32 & 84.38 & 66.41 & 2.63 & 1.67 & 2.30 & 1.17 & -2.58 & 0.0049 \\
    \hline
    51 & road, car & 21 & 100.00 & 47.65 & 5.32 & 1.52 & 5.62 & 0.00 & -6.03 & \textless .00001 \\
    54 & skyscraper & 39 & 100.00 & 71.78 & 4.14 & 1.61 & 4.08 & 1.12 & -7.60 & \textless .00001 \\
    56 & flusher, soap\_dish & 53 & 92.45 & 64.29 & 3.47 & 1.48 & 3.08 & 0.86 & -6.47 & \textless .00001 \\
    57 & shower\_stall, screen\_door & 34 & 97.06 & 32.31 & 2.60 & 0.61 & 2.53 & 0.00 & -7.55 & \textless .00001 \\
    63 & edifice, skyscraper & 45 & 88.89 & 48.38 & 2.41 & 0.83 & 2.36 & 0.00 & -6.73 & \textless .00001 \\
    \hline
\end{tabular}}
\end{table}

\begin{table}
\caption{CLIP-Dissect -- Evaluation details as discussed in Section~\ref{subsubsec:eval}.
Images: Number of images used for evaluation.
\# Activations: (targ(et)): Percentage of target images activating the neuron (i.e.,
activation at least 80\% of this neuron's activation maximum);
(\mbox{non-t}):
Same for all
other images used in the evaluation. Mean/Median (targ(et)/non-t(arget)): Mean/median activation value for
target and non-target images, respectively.}
\label{tab:evaluationClip}
\centering
\resizebox{\columnwidth}{!}{%
\begin{tabular}{clc|rr|rr|rr|rr}
     \multicolumn{1}{l}{Neuron} & \multicolumn{1}{l}{Label(s)} & \multicolumn{1}{c}{Images} & \multicolumn{2}{c}{\# Activations (\%)} & \multicolumn{2}{c}{Mean} & \multicolumn{2}{c}{Median} & \multicolumn{1}{c}{z-score} & \multicolumn{1}{c}{p-value}\\
     \hline
     &  & & targ & non-t & targ & non-t & targ & non-t &  & \\
    \hline
    3 & dresser & 43 & 93.02 & 64.61 & 2.59 & 1.42 & 2.62 & 0.68 & 5.01 & \textless{}0.0001 \\
    7 & bathroom & 46 & 89.47 & 41.56 & 2.02 & 1.01 & 2.15 & 0.00 & 5.45 & \textless{}0.0001 \\
    18 & dining & 36 & 94.87 & 76.82 & 3.01 & 1.85 & 3.11 & 1.44 & 4.52 & \textless{}0.0001 \\
    33 & bathroom & 38 & 71.05 & 34.02 & 1.28 & 0.47 & 0.95 & 0.00 & 4.91 & \textless{}0.0001 \\
    38 & bathroom & 38 & 84.21 & 31.71 & 1.79 & 0.54 & 1.83 & 0.00 & 7.14 & \textless{}0.0001 \\
    43 & highways & 32 & 100.00 & 63.87 & 7.00 & 3.14 & 6.39 & 2.64 & 6.17 & \textless{}0.0001 \\
    49 & bedroom & 40 & 97.50 & 55.77 & 3.48 & 1.63 & 3.43 & 0.63 & 6.05 & \textless{}0.0001 \\
    50 & bedroom & 40 & 97.50 & 63.21 & 4.56 & 1.30 & 4.60 & 0.66 & 8.70 & \textless{}0.0001 \\

    \hline
\end{tabular}}
\end{table}

\begin{table}
\caption{GPT-4 -- Evaluation details as discussed in Section~\ref{subsubsec:eval}.
Images: Number of images used for evaluation.
\# Activations: (targ(et)): Percentage of target images activating the neuron (i.e.,
activation at least 80\% of this neuron's activation maximum);
(\mbox{non-t}):
Same for all
other images used in the evaluation. Mean/Median (targ(et)/non-t(arget)): Mean/median activation value for
target and non-target images, respectively.}
\label{tab:evaluationGPT}
\centering
\resizebox{\columnwidth}{!}{%
\begin{tabular}{clc|rr|rr|rr|rr}
     \multicolumn{1}{l}{Neuron} & \multicolumn{1}{l}{Label(s)} & \multicolumn{1}{c}{Images} & \multicolumn{2}{c}{\# Activations (\%)} & \multicolumn{2}{c}{Mean} & \multicolumn{2}{c}{Median} & \multicolumn{1}{c}{z-score} & \multicolumn{1}{c}{p-value}\\
     \hline
     &  & & targ & non-t & targ & non-t & targ & non-t &  & \\
    \hline
1 & Street Scene & 42 & 90.50 & 30.40 & 3.80 & 0.70 & 4.20 & 0.00 & -9.62 & \textless 0.0001 \\
3 & Bedroom & 42 & 97.60 & 63.40 & 4.70 & 1.20 & 4.90 & 0.70 & -9.05 & \textless 0.0001 \\
6 & Kitchen & 43 & 83.70 & 52.00 & 2.40 & 1.00 & 2.00 & 0.10 & -5.06 & \textless 0.0001 \\
8 & Bathroom & 41 & 100.00 & 44.10 & 4.10 & 1.00 & 4.10 & 0.00 & -9.57 & \textless 0.0001 \\
14 & Living Room & 41 & 78.00 & 67.50 & 1.40 & 1.30 & 1.20 & 0.90 & -0.77 & 0.4413 \\
\hline
17 & Dining Room & 40 & 97.50 & 45.90 & 2.20 & 0.60 & 2.50 & 0.00 & -8.29 &\textless 0.0001 \\
18 & Outdoor Scenery & 41 & 100.00 & 76.10 & 2.30 & 1.50 & 2.20 & 1.20 & -3.96 & \textless 0.0001 \\
22 & Street Scene & 42 & 90.50 & 50.10 & 3.00 & 1.40 & 3.30 & 0.00 & -5.95 & \textless 0.0001 \\
23 & Street Scene & 42 & 85.70 & 20.70 & 2.40 & 0.30 & 2.10 & 0.00 & -10.83 & \textless 0.0001 \\
29 & Bathroom & 41 & 90.20 & 68.40 & 2.60 & 1.50 & 2.40 & 1.00 & -4.05 & \textless 0.0001 \\
\hline
30 & Kitchen & 43 & 86.00 & 38.60 & 2.60 & 0.80 & 2.70 & 0.00 & -7.22 & \textless 0.0001 \\
31 & Urban Street Scene & 41 & 80.50 & 65.70 & 1.80 & 1.30 & 1.70 & 0.90 & -2.4 & 0.164 \\
36 & Bathroom & 41 & 100.00 & 61.30 & 3.10 & 1.20 & 2.80 & 0.60 & -7.48 & \textless 0.0001 \\
38 & Living Room & 41 & 92.70 & 54.30 & 2.00 & 1.00 & 2.20 & 0.30 & -5.53 & \textless 0.0001 \\
39 & Bicycle & 39 & 84.60 & 47.40 & 2.10 & 0.90 & 2.40 & 0.00 & -5.64 & \textless 0.0001 \\
\hline
41 & Living Room & 41 & 97.60 & 42.00 & 2.60 & 0.60 & 2.30 & 0.00 & -9.31 & \textless 0.0001 \\
43 & Outdoor Urban Scene & 41 & 92.70 & 56.30 & 4.10 & 2.40 & 4.30 & 1.00 & -4.42 & \textless 0.0001 \\
44 & Kitchen Scene & 42 & 81.00 & 43.40 & 2.30 & 1.00 & 2.10 & 0.00 & -5.43 & \textless 0.0001 \\
48 & Urban Street Scene & 41 & 100.00 & 52.60 & 4.90 & 2.30 & 4.80 & 0.40 & -6.03 & \textless 0.0001 \\
49 & Bedroom & 42 & 95.20 & 35.00 & 3.80 & 0.70 & 4.00 & 0.00 & -10.31 & \textless 0.0001 \\
\hline
50 & Living Room & 41 & 97.60 & 63.90 & 3.00 & 1.20 & 2.60 & 0.60 & -6.78 & \textless 0.0001 \\
51 & Street Scene & 42 & 95.20 & 42.90 & 5.70 & 1.50 & 6.10 & 0.00 & -9.05 & \textless 0.0001 \\
56 & Toilet Brush & 42 & 97.60 & 34.60 & 3.60 & 0.70 & 3.60 & 0.00 & -10.48 & \textless 0.0001 \\
57 & Bathroom Interior & 41 & 92.70 & 40.50 & 3.00 & 0.80 & 2.90 & 0.00 & -8.35 & \textless 0.0001 \\
59 & Urban Street Scene & 41 & 82.90 & 26.30 & 2.70 & 0.50 & 2.50 & 0.00 & -9.06 & \textless 0.0001 \\
62 & Dining Room & 40 & 90.00 & 43.90 & 3.30 & 0.80 & 3.70 & 0.00 & -8.64 & \textless 0.0001 \\
63 & Cityscape & 39 & 97.40 & 48.50 & 2.80 & 0.70 & 2.40 & 0.00 & -8.76 & \textless 0.0001 \\
    \hline
\end{tabular}}
\end{table}

\begin{table}
\caption{Concept Accuracy in Hidden Layer Activation Space of Concepts extracted using Concept Induction, where the training and test dataset is GoogleImage Dataset.}
\label{tab:concept_accuracy_long}
\centering
\resizebox{.70\columnwidth}{!}{
\begin{footnotesize}
\begin{tabular}{clrrrr}
Concept Name & Method & Train Accuracy & Test Accuracy\\
\hline
        Air Conditioner&  CAR&  0.8994& 0.8415\\ 
         Air Conditioner&  CAV&  0.811& 0.8659\\ 
         Baseboard&  CAR&  0.875& 0.8717\\ 
         Baseboard&  CAV&  0.8846& 0.9102\\ 
         Bushes&  CAR&  0.9150& 0.9487\\ 
         Bushes&  CAV&  0.9477& 0.9743\\ 
         Car&  CAR&  0.9464& 0.9571\\ 
         Car&  CAV&  0.925& 0.9429\\
         Coffee Table&  CAR&  0.9047& 0.9523\\ 
         Coffee Table&  CAV&  0.8988& 0.9166\\
         Cross Walk&  CAR&  0.9166& 0.9468\\ 
         Cross Walk&  CAV&  0.9247& 0.9361\\
         Dishcloth&  CAR&  0.9055& 0.9375\\ 
         Dishcloth&  CAV&  0.9685& 0.9531\\ 
         Doorcase&  CAR&  0.8936& 0.8611\\ 
         Doorcase&  CAV&  0.8581& 0.8194\\ 
         Edifice&  CAR&  0.9487& 0.9642\\  
         Edifice&  CAV&  0.9548& 0.9523\\ 
         Fire Hydrant&  CAR&  0.9171& 0.9625\\ 
         Fire Hydrant&  CAV&  0.9171& 0.925\\  
         Footboard&  CAR&  0.9268& 0.9519\\ 
         Footboard&  CAV&  0.9585&	0.9423\\
         Flusher&  CAR&  0.8722& 0.8285\\ 
         Flusher&  CAV&  0.9014& 0.9285\\ 
         Fluorescent Tube&  CAR&  0.9006& 0.9625\\ 
         Fluorescent Tube&  CAV&  0.9358& 0.9125\\ 
         Manhole&  CAR&  0.9349& 0.8953\\ 
         Manhole&  CAV&  0.9349& 0.9302\\
         Nuts&  CAR&  0.9223& 0.9134\\ 
         Nuts&  CAV&  0.9417& 0.9230\\
         Paper Towels&  CAR&  0.9021& 0.9166\\ 
         Paper Towels&  CAV&  0.9239& 0.9166\\ 
         Pipage&  CAR&  0.84239& 0.7826\\ 
         Pipage&  CAV&  0.7826& 0.7391\\ 
         Posters&  CAR&  0.8806& 0.9230\\ 
         Posters&  CAV&  0.8806& 0.9230\\ 
         Pylon&  CAR&  0.8397& 0.8125\\ 
         Pylon&  CAV&  0.8205& 0.8375\\ 
         River&  CAR&  0.9430& 0.925\\ 
         River&  CAV&  0.9399& 0.925\\  
         Slope&  CAR&  0.8705& 0.8714\\ 
         Slope&  CAV&  0.9208& 0.8857\\ 
         Sculpture&  CAR&  0.8242& 0.8333\\ 
         Sculpture&  CAV&  0.8788& 0.8571\\ 
         Screen Door&  CAR&  0.9076& 0.9375\\ 
         Screen Door&  CAV&  0.9235& 0.925\\ 
         Spatula&  CAR&  0.9017& 0.9431\\ 
         Soap Dispenser&  CAR&  0.88& 0.9375\\ 
         Soap Dispenser&  CAV&  0.916& 0.9531\\ 
         Spatula&  CAV&  0.9219& 0.9204\\  
         Toaster&  CAR&  0.927& 0.9714\\ 
         Toaster&  CAV&  0.9197& 0.9736\\ 
         Wardrobe&  CAR&  0.9375& 0.95\\ 
         Wardrobe&  CAV&  0.9188& 0.9125\\ 
\hline
\end{tabular}
\end{footnotesize}}
\end{table}

\begin{table}
\centering
\resizebox{.70\columnwidth}{!}{
\begin{footnotesize}
\begin{tabular}{clrrrr}
Concept Name & Method & Train Accuracy & Test Accuracy\\
\hline
     
         Body&  CAR&  0.9035& 0.8857\\ 
         Body&  CAV&  0.8642& 0.9\\ 
         Casserole&  CAR&  0.9458& 0.9375\\ 
         Casserole&  CAV&  0.9808& 0.975\\ 
         Central Reservation&  CAR&  0.8694& 0.9\\ 
         Central Reservation&  CAV&  0.8917& 0.9\\
         Chain&  CAR&  0.9556& 0.9677\\ 
         Chain&  CAV&  0.9637& 0.9677\\ 
         
         Cistern&  CAR&  0.8734& 0.8375\\ 
         Cistern&  CAV&  0.8449& 0.8875\\ 
         Crapper&  CAR&  0.8516& 0.8043\\ 
         Crapper&  CAV&  0.8571& 0.8695\\
         Dish Rack&  CAR&  0.9375& 0.9583\\ 
         Dish Rack&  CAV&  0.9843& 0.9375\\ 
         Fire Escape&  CAR&  0.8950& 0.9146\\ 
         Fire Escape&  CAV&  0.9104& 0.8902\\ 
         Flooring&  CAR&  0.8841& 0.9166\\ 
         Flooring&  CAV&  0.8871& 0.9047\\ 
         Go Cart&  CAR&  0.9378& 0.9512\\ 
         Go Cart&  CAV&  0.9254& 0.9390\\ 
         Jar&  CAR&  0.9059& 0.9333\\ 
         Jar&  CAV&  0.9572& 0.9666\\
         Left Foot&  CAR&  0.8734& 0.8658\\ 
         Left Foot&  CAV&  0.8703& 0.8536\\ 
         Lid&  CAR&  0.8622& 0.9047\\ 
         Lid&  CAV&  0.8712& 0.8809\\ 
         
         Mouth&  CAR&  0.8963& 0.9268\\ 
         Mouth&  CAV&  0.9481& 0.9512\\ 
         Night Table&  CAR&  0.8917& 0.875\\ 
         Night Table&  CAV&  0.9235& 0.8875\\
         Open Fireplace&  CAR&  0.9129& 0.9222\\
         Open Fireplace&  CAV&  0.9101& 0.9333\\ 
         Ornament&  CAR&  0.8910& 0.9375\\ 
         Ornament&  CAV&  0.9198& 0.9625\\ 
         Pillar&  CAR&  0.8372& 0.8837\\ 
         Pillar&  CAV&  0.7732& 0.8372\\
         Plank&  CAR&  0.8719& 0.9523\\ 
         Plank&  CAV&  0.9146& 0.9047\\
         Road&  CAR&  0.9221& 0.9642\\ 
         Road&  CAV&  0.9461& 0.9404\\ 
         Rocking Horse&  CAR&  0.9173& 0.9310\\ 
         Rocking Horse&  CAV&  0.9347& 0.9655\\
         Saucepan&  CAR&  0.9561& 0.9827\\ 
         Saucepan&  CAV&  1& 0.9827\\
         Sideboard&  CAR&  0.91& 0.94\\ 
         Sideboard&  CAV&  0.965& 0.92\\
         Shower Stall&  CAR&  0.9409& 0.9722\\ 
         Shower Stall&  CAV&  0.9652& 0.9583\\
         Skyscraper&  CAR&  0.9455&	0.9743\\ 
         Skyscraper&  CAV&  0.9615& 0.9743\\ 
         Slipper&  CAR&  0.9262& 0.9456\\ 
         Slipper&  CAV&  0.9617& 0.9565\\
         Soap Dish&  CAR&  0.8733& 0.8589\\ 
         Soap Dish&  CAV&  0.8474& 0.8589\\ 
         Stem&  CAR&  0.8834& 0.8676\\ 
         Stem&  CAV&  0.8383& 0.8382\\
         Tap&  CAR&  0.8198& 0.8536\\ 
         Tap&  CAV&  0.8354& 0.8902\\ 
          
\hline
\end{tabular}
\end{footnotesize}}
\end{table}

\begin{table}
\centering
\resizebox{.70\columnwidth}{!}{
\begin{footnotesize}
\begin{tabular}{clrrrr}
Concept Name & Method & Train Accuracy & Test Accuracy\\
\hline
        Building&  CAR&  0.9085& 0.9404\\ 
         Building&  CAV&  0.8262& 0.8690\\ 
         Dishrag&  CAR&  0.8603& 0.9285\\ 
         Dishrag&  CAV&  0.9144& 0.9464\\ 
         Left Arm&  CAR&  0.8549& 0.8536\\ 
         Left Arm&  CAV&  0.8858& 0.8658\\ 
         Letter Box&  CAR&  0.8901& 0.8636\\ 
         Letter Box&  CAV&  0.875& 0.9242\\
         Mountain&  CAR&  0.9426& 0.95\\ 
         Mountain&  CAV&  0.9745& 0.9625\\
         River Water&  CAR&  0.9554& 0.9375\\ 
         River Water&  CAV&  0.9617& 0.9375\\
         Rocker&  CAR&  0.8953& 0.9545\\ 
         Rocker&  CAV&  0.9457& 0.8939x\\
         Side Rail&  CAR&  0.9054& 0.9459\\ 
         Side Rail&  CAV&  0.8986& 0.9054\\
         Stretcher&  CAR&  0.89375& 0.9375\\ 
         Stretcher&  CAV&  0.9312&	0.9375\\
         Tank Lid&  CAR&  0.8947& 0.8846\\ 
         Tank Lid&  CAV&  0.8848& 0.8717\\
         Teapot&  CAR&  0.9365& 0.9411\\ 
         Teapot&  CAV&  0.9552& 0.9779\\ 
         Toothbrush&  CAR&  0.9198& 0.9125\\ 
         Toothbrush&  CAV&  0.9198& 0.9\\ 
         Utensils Canister&  CAR&  0.9262&	0.925\\ 
         Utensils Canister&  CAV&  0.9487&	0.9375\\

\hline
\end{tabular}
\end{footnotesize}}
\end{table}

\begin{table}
\caption{Concept Accuracy in Hidden Layer Activation Space of Combinations of Concepts extracted using Concept Induction, where the training and test dataset is GoogleImage Dataset.}
\label{tab:concept_accuracy_combined}
\centering
\resizebox{.80\columnwidth}{!}{
\begin{footnotesize}
\begin{tabular}{clrrrr}
Concept Name & Method & Train Accuracy & Test Accuracy\\
\hline
    Baseboard and Dishrag&  CAR&  0.8694& 0.8519\\ 
         Baseboard and Dishrag&  CAV&  0.8611& 0.8519\\ 
         Cistern and Doorcase&  CAR&  0.8971& 0.9038\\ 
         Cistern and Doorcase&  CAV&  0.9207& 0.9038\\ 
         Dishcloth and Toaster&  CAR&  0.9190& 0.8889\\ 
         Dishcloth and Toaster&  CAV&  0.9231& 0.9259\\ 
         Edifice and Skyscraper&  CAR&  0.8991& 0.8778\\ 
         Edifice and Skyscraper&  CAV&  0.8733& 0.8667\\ 
         Flooring and Fluorescent Tube&  CAR&  0.8113& 0.7931\\ 
         Flooring and Fluorescent Tube&  CAV&  0.8652& 0.8442\\ 
         Flusher and Soap Dish&  CAR&  0.8649& 0.8679\\ 
         Flusher and Soap Dish&  CAV&  0.8538& 0.8302\\ 
         Footboard and Chain&  CAR&  0.9047& 0.8906\\ 
         Footboard and Chain&  CAV&  0.9246& 0.8906\\ 
         Left Foot and Mouth&  CAR&  0.9588& 0.9464\\ 
         Left Foot and Mouth&  CAV&  0.9311& 0.9107\\ 
         Letter Box and Go Cart&  CAR&  0.8800& 0.8750\\ 
         Letter Box and Go Cart&  CAV&  0.8680& 0.8437\\ 
         Lid and Soap Dispenser&  CAR&  0.9237& 0.9394\\ 
         Lid and Soap Dispenser&  CAV&  0.9722& 0.9697\\ 
         Manhole and Left Arm&   CAR&  0.8647& 0.8636\\ 
         Manhole and Left Arm&   CAV&  0.8388& 0.8181\\ 
         Mountain and Bushes&   CAR&  0.9933& 0.9815\\ 
         Mountain and Bushes&   CAV&  0.9678& 0.9444\\ 
         Open Fireplace and Coffee Table&  CAR&  0.9655& 0.9838\\  
         Open Fireplace and Coffee Table&  CAV&  0.9508& 0.9354\\ 
         Ornament and Saucepan&  CAR&  0.8872& 0.8846\\  
         Ornament and Saucepan&  CAV&  0.8363& 0.8269\\ 
         Paper Towel and Jar&  CAR&  0.9133& 0.9090\\  
         Paper Towel and Jar&  CAV&  0.8719& 0.8409\\ 
\hline
\end{tabular}
\end{footnotesize}}
\end{table}

\begin{table}
\label{tab:concept_accuracy_combined_2}
\centering
\resizebox{.80\columnwidth}{!}{
\begin{footnotesize}
\begin{tabular}{clrrrr}
Concept Name & Method & Train Accuracy & Test Accuracy\\
\hline
            Pillar and Stretcher&  CAR&  0.7847& 0.7667\\  
         Pillar and Stretcher&  CAV&  0.8290& 0.8000\\ 
         Plank and Casserole&  CAR&  0.8836& 0.8750\\  
         Plank and Casserole&  CAV&  0.8719& 0.8750\\ 
         Pylon and Posters&  CAR&  0.8605& 0.8461\\ 
         Pylon and Posters&  CAV&  0.8509& 0.8269\\ 
         Road and Car&   CAR&  0.9661& 0.9524\\ 
         Road and Car&   CAV&  0.9104& 0.8909\\ 
         Rocking Horse and Rocker&   CAR&  0.9844& 0.9773\\ 
         Rocking Horse and Rocker&   CAV&  0.9651& 0.9545\\ 
         Saucepan and Dishrack&  CAR&  0.8916& 0.8870\\  
         Saucepan and Dishrack&  CAV&  0.9416& 0.9355\\ 
         Sculpture and Siderail&  CAR&  0.7635& 0.7500\\  
         Sculpture and Siderail&  CAR&  0.8201& 0.8167\\  
         Shower Stall and Cistern&  CAR&  0.9787& 0.9571\\  
         Shower Stall and Cistern&  CAV&  0.9438& 0.9571\\ 
         Shower Stall and Screen Door&  CAR&  0.9706& 0.9549\\  
         Shower Stall and Screen Door&  CAV&  0.9853& 0.9849\\ 
         Skyscaper and River&  CAR&  0.9596& 0.9483\\  
         Skyscaper and River&  CAV&  0.9471& 0.9310\\ 
         Spatula and Nuts&  CAR&  0.8356& 0.8125\\  
         Spatula and Nuts&  CAV&  0.9245& 0.9063\\ 
         Tap and Crapper&  CAR&  0.8861& 0.8913\\  
         Tap and Crapper&  CAV&  0.8695& 0.8913\\ 
         Teapot and Saucepan&  CAR&  0.9881& 0.9629\\  
         Teapot and Saucepan&  CAV&  0.9637& 0.9444\\ 
         Toothbrush and Pipage&  CAR&  0.7971& 0.7857\\  
         Toothbrush and Pipage&  CAV&  0.8373& 0.8214\\ 
         Utensils Canister and Body&  CAR&  0.9579& 0.9464\\  
         Utensils Canister and Body&  CAV&  0.9299& 0.9107\\ 
         Wardrobe and Air Conditioning&  CAR&  0.9309& 0.9286\\  
         Wardrobe and Air Conditioning&  CAV&  0.9118& 0.9107\\ 
\hline
\end{tabular}
\end{footnotesize}}
\end{table}

\begin{table}
\centering
\caption{Concept Accuracy in Hidden Layer Activation Space of Concepts extracted using CLIP-Dissect.}
\label{tab:concept_accuracy_clip_dissect}
\resizebox{.70\columnwidth}{!}{
\begin{footnotesize}
\begin{tabular}{clrrrr}
Concept Name & Method & Train Accuracy & Test Accuracy\\
\hline
        Bathroom&  CAR&  0.9700& 0.9474\\
         Bathroom&  CAV&  0.9400& 0.9474\\
        Bed&  CAR&  0.9587& 0.9500\\
        Bed&  CAV&  0.9437& 0.9125\\
        Bedroom&  CAR&  0.9167& 0.9167\\
        Bedroom&  CAV&  0.9137& 0.9048\\
        Buildings&  CAR&  0.9321& 0.9230\\
        Buildings&  CAV&  0.8990& 0.8974\\
        Dallas&  CAR&  0.9447& 0.9318\\
         Dallas&  CAV&  0.9750& 0.9545\\
        Dining&  CAR&  0.9294& 0.9125\\
         Dining&  CAV&  0.8907& 0.9000\\
         Dresser&  CAR&  0.9762& 0.9625\\
         Dresser&  CAV&  0.9650& 0.9500\\
         File&  CAR&  0.9837& 0.9750\\
         File&  CAV&  0.9681& 0.9500\\
         Furnished&  CAR&  0.8843& 0.8875\\
         Furnished&  CAV&  0.8762& 0.8625\\
         Highways&  CAR&  0.9396& 0.9375\\
         Highways&  CAV&  0.9679& 0.9531\\
         Interstate&  CAR&  0.9293& 0.9268\\
         Interstate&  CAV&  0.8593& 0.8536\\
         Kitchen&  CAR&  9848& 0.9743\\
         Kitchen&  CAV&  0.9590& 0.9487\\
         Legislature&  CAR&  0.9149& 0.9000\\
         Legislature&  CAV&  0.9156& 0.9000\\
        Microwave&  CAR&  0.9803& 0.9807\\
        Microwave&  CAV&  0.9873& 0.9807\\
        Mississauga&  CAR&  0.9041& 0.9054\\
         Mississauga&  CAV&  0.9467& 0.9324\\
         Municipal&  CAR&  0.8679& 0.8461\\
         Municipal&  CAV&  0.9298& 0.9102\\
         Restaurants&  CAR&  0.9850& 0.9722\\
         Restaurants&  CAV&  0.9692& 0.9583\\
         Road&  CAR&  0.9362& 0.9250\\
         Road&  CAV&  0.9387& 0.9250\\
         Room&  CAR&  0.8653& 0.8125\\
         Room&  CAV&  0.8273& 0.8250\\
         Roundtable&  CAR&  0.9405& 0.9473\\
         Roundtable&  CAV&  0.9136& 0.8947\\
         Street&  CAR&  0.9830& 0.9722\\
         Street&  CAV&  0.9347& 0.9167\\
         Valencia&  CAR&  0.8735& 0.8625\\
         Valencia&  CAV&  0.8781& 0.875\\
\hline
\end{tabular}
\end{footnotesize}}
\end{table}

\begin{table}
\centering
\caption{Concept Accuracy in Hidden Layer Activation Space of Concepts extracted using GPT-4.}
\label{tab:concept_accuracy_gpt4}
\resizebox{.70\columnwidth}{!}{
\begin{footnotesize}
\begin{tabular}{clrrrr}
Concept Name & Method & Train Accuracy & Test Accuracy\\
\hline
    Bedroom&  CAR&  0.9851& 0.9761\\ 
         Bedroom&  CAV&  0.9660& 0.9523\\ 
         Bathroom Interior&  CAR&  0.9273& 0.9146\\
         Bathroom Interior&  CAV&  0.9241& 0.9268\\

         Bathroom&  CAR&  0.9176& 0.9024\\
         Bathroom&  CAV&  0.9068& 0.8902\\
         Bicycle&  CAR&  0.9787& 0.9615\\
         Bicycle&  CAV&  0.9887& 0.9871\\
         Cityscape&  CAR&  0.9438& 0.9358\\
         Cityscape&  CAV&  0.9894& 0.9743\\
         Classroom&  CAR&  0.8981& 0.8780\\
         Classroom&  CAV&  0.9012& 0.8536\\
         Dining Room&  CAR&  0.9256& 0.9125\\
         Dining Room&  CAV&  0.8942& 0.8875\\
         Eyeglasses&  CAR&  0.9813& 0.9883\\
         Eyeglasses&  CAV&  0.9883& 0.9883\\
         Home Interior&  CAR&  0.8515& 0.8452\\
         Home Interior&  CAV&  0.8363& 0.8214\\
         Indoor Home Setting&  CAR&  0.6713& 0.6785\\
         Indoor Home Setting&  CAV&  0.6890& 0.6666\\
         Indoor Home Decor&  CAR&  0.8428& 0.8333\\
         Indoor Home Decor&  CAV&  0.8418& 0.8222\\
         Kitchen Scene&  CAR&  0.8562& 0.8571\\
         Kitchen Scene&  CAV&  0.8022& 0.7976\\
         Kitchen&  CAR&  0.9122& 0.9302\\
         Kitchen&  CAV&  0.9122& 0.9186\\
         Living Room&  CAR&  0.8963& 0.8658\\
         Living Room&  CAV&  0.8658& 0.8414\\
         Outdoor Scenery&  CAR&  0.9135& 0.9024\\
         Outdoor Scenery&  CAV&  0.9054& 0.9024\\
         Outdoor Urban Scene&  CAR&  0.8343& 0.8170\\
         Outdoor Urban Scene&  CAV&  0.7650& 0.7317\\
         Street Scene&  CAR&  0.8819& 0.8809\\
         Street Scene&  CAV&  0.8568& 0.8690\\
         Toilet Brush&  CAR&  0.9815& 0.9761\\
         Toilet Brush&  CAV&  0.9727& 0.9642\\
         Urban Landscape&  CAR&  0.8665& 0.8636\\
         Urban Landscape&  CAV&  0.8922& 0.8863\\
         Urban Transportation&  CAR&  0.8412& 0.8414\\
         Urban Transportation&  CAV&  0.8251& 0.8414\\
         Urban Street Scene&  CAR&  0.9140& 0.9024\\
         Urban Street Scene&  CAV&  0.8757& 0.8658\\
\hline
\end{tabular}
\end{footnotesize}}
\end{table}

\begin{table}
\caption{Summary of Concept Activation Analysis results of Concept Induction, CLIP-Dissect, and GPT-4 using Mann-Whitney U test}
\label{tab:mwuconceptactivation}
\centering
\resizebox{.80\columnwidth}{!}{
\begin{footnotesize}
\begin{tabular}{l||r|r||r|r}
     \multicolumn{1}{l}{Method} & \multicolumn{2}{c}{CAV} & \multicolumn{2}{c}{CAR}\\
     \hline
     & z-score & p-value & z-score & p-value\\
    \hline
    Concept Induction x CLIP-Dissect & 0.1252 & 0.9004 & -0.8717 & 0.3834 \\
    CLIP-Dissect x GPT-4 & 1.7494 & 0.0801 & \textbf{1.9680} & \textbf{0.0488} \\
    Concept Induction x GPT-4 & \textbf{2.1560} & \textbf{0.0308} & 1.7792 & 0.0751 \\
    \hline
\end{tabular}
\end{footnotesize}}
\end{table}

\begin{table}
\caption{Summary of Concept Activation Analysis results of Concept Induction, CLIP-Dissect, and GPT-4 using Mean and Median of their respective test accuracies}
\label{tab:meanandmedianconceptactivation}
\centering
\resizebox{.90\columnwidth}{!}{
\begin{footnotesize}
\begin{tabular}{l||r|r|r||r|r|r}
     \multicolumn{1}{l}{Method} & \multicolumn{3}{c}{CAV} & \multicolumn{3}{c}{CAR}\\
     \hline
     & Mean & Median & Std. Deviation & Mean & Median & Std. Deviation\\
    \hline
    Concept Induction& 0.9154 & 0.9230 & 0.0449 & 0.9150 & 0.9310 & 0.0465 \\
    CLIP-Dissect& 0.9160 & 0.9146 & 0.0389 & 0.9259 & 0.9293 &  0.0443 \\
    GPT-4 & 0.8757 &  0.8863 & 0.0817 & 0.8887 & 0.9024 & 0.0690 \\
    \hline
\end{tabular}
\end{footnotesize}}
\end{table}

\section{Discussion}
\label{sec:discussion}

From the results of the statistical evaluation from each method, based on the percentage of target activation, if we were to categorize all confirmed concepts into three regions: high-relevance concepts (90-100\%), medium-relevance concepts (80-89\%), and low-relevance concepts ($< 80\%$), Table~\ref{tab:statevalpercentageofconcepts} illustrates that Concept Induction produces a notably larger number of high-relevance concepts compared to CLIP-Dissect and GPT-4. 

In Tables~\ref{tab:evaluationClip},~\ref{tab:evaluationGPT}, we present 8 and 27 statistically confirmed concepts, respectively. However, upon closer examination, it becomes evident that some concepts are duplicated across the tables. If we disregard these duplicates, Table~\ref{tab:evaluationClip} and ~\ref{tab:evaluationGPT} would reveal 5 and 14 confirmed concepts, respectively. Eliminating duplicate concepts is essential for accurately assessing the confirmed concepts' diversity and significance, as having duplicate concepts means we are essentially assessing the same concept multiple times. This redundancy may not add any new insights and can lead to misleading results.

From the results of Concept Activation Analysis, based on the concepts' test accuracy, if we are to bin all the concepts into 3 regions: high relevance concepts 90-100\%, medium relevance concepts 80-89\%, and low relevance concepts $< 80\%$, Table~\ref{tab:conceptactivationcountofconcepts} illustrates that Concept Induction produces a notably larger number of high-relevance concepts compared to CLIP-Dissect and GPT-4, while the latter two exhibit similar counts. 

This disparity highlights Concept Induction's reliance on a rich background knowledge base, necessitating additional preprocessing but providing additional value. We argue that the candidate concept pool of 20K English vocabulary words or off-the-shelf GPT-4 do not work as one size that fits all. Concept Induction's ability to generate extensive high-relevance concepts underscores the importance of well-engineered background knowledge. That is not to say that CLIP-Dissect/GPT-4 do not provide any usefulness, there exists a trade-off of time. If the application does not require a more comprehensive set of concept-based explanations, then CLIP-Dissect/GPT-4 may appear as a useful solution especially when time is of the essence. That being said, if the application requires detailed analysis of concept-based explanations, then we argue that it is important to prepare a background knowledge and leverage Concept Induction to gain detailed insights. 

In techniques like CLIP-Dissect and GPT-4, it is unclear how to meticulously craft the pool of candidate concepts. Employing a background knowledge base it is possible to make use of relationships among concepts. Concept Induction facilitates deductive reasoning utilizing this background knowledge, inherently offering transparency compared to the opaque approach of CLIP-Dissect and GPT-4 in concept extraction. This affords users flexibility in shaping the candidate concept pool, potentially serving as a valuable tool in specialized applications where a general concept pool is of limited relevance. 

Our thorough analysis emphasizes the importance of exploring the selection criteria for candidates in an Explainable AI approach. While it's crucial to investigate methods that assess the relevance of concepts in hidden layer computations within a given candidate pool, it is equally, if not more, vital to thoughtfully design this pool. Neglecting this aspect could result in overlooking crucial concepts essential for gaining insights into hidden layer computations. Our approach offers a means to integrate rich background knowledge and extract meaningful concepts from it.

One drawback of utilizing Concept Induction (and GPT-4) is its dependency on labeled image data, specifically image annotations, which serve as data points in the background knowledge. In contrast, CLIP-Dissect operates without the need for labels and can function with any provided set of images. 
We view this as a trade-off that must be carefully considered based on the application scenario. If the application is broad and does not demand a meticulous design of candidate concepts, then employing approaches like CLIP-Dissect can be advantageous. Conversely, for applications that are focused or specialized, CLIP-Dissect may only provide broadly relevant concepts.

While our thorough evaluation provides valuable insights into how our method performs compared to existing techniques, our focus has been primarily on assessing the effectiveness of Concept Induction within the confines of Convolutional Neural Network architecture using ADE20K Image data. Nevertheless, it is imperative to investigate its suitability across different architectures and with diverse datasets. Given the model-agnostic nature of our approach, our results suggest its potential applicability across a range of neural network architectures, datasets, and modalities. Another avenue for exploration involves examining background knowledge of varying scales and assessing its impact on extracting highly relevant concepts. While we utilized a Wiki Concept Hierarchy comprising 20 million concepts, it would be intriguing to observe the outcomes of our approach when powered by a domain-specific Knowledge Graph in specialized domains such as Medical Diagnosis.

\begin{table}
\caption{Count of statistically confirmed Concepts from each method (Table~\ref{tab:evaluationCI},~\ref{tab:evaluationClip},~\ref{tab:evaluationGPT}) such that their percentage of target activation is binned into 3 regions based on their degree of relevance.}
\label{tab:statevalpercentageofconcepts}
\centering

\begin{footnotesize}
\begin{tabular}{l|l|l|l}
Method & 90-100\% & 80-89\% & \textless{}80\% \\
\hline
Concept Induction & 14 & 6 & 0 \\
GPT-4 & 10 & 4 & 0 \\
CLIP-Dissect & 4 & 1 & 0 \\
\end{tabular}
\end{footnotesize}
\end{table}

\begin{table}
\caption{Count of Concepts from each method such that their concept classifier's test accuracies are binned into 3 regions based on their degree of relevance.}
\label{tab:conceptactivationcountofconcepts}
\centering

\begin{footnotesize}
\begin{tabular}{l|l|l|l}
Method & 90-100\% & 80-89\% & \textless{}80\% \\
\hline
Concept Induction & 46 & 22 & 1 \\
CLIP-Dissect & 17 & 5 & 0 \\
GPT-4 & 11 & 9 & 1 \\

\end{tabular}
\end{footnotesize}
\end{table}


\section{Conclusion}
\label{sec:conclusion}
We have shown that our use of Concept Induction and large-scale ontological background knowledge leads to meaningful labeling of hidden neuron activations, confirmed by statistical analysis, which assesses the activation of neurons for both target and non-target concepts. This enables us to identify concepts that trigger the most pronounced responses from the neurons and discern meaningful patterns and associations between concepts and neuron activations, thereby enhancing the robustness and validity of our labeling methodology. Additionally, we employ Concept Activation Analysis to measure the degree of relevance of each concept across the entire dense layer activation space. This combined approach allows us to delve into the inner workings of hidden layer computations, providing a comprehensive understanding of how individual neurons function within the broader context of the dense layer's operation.  To the best of our knowledge, this approach is new, and in particular the use of large-scale background knowledge for this purpose – meaning that label categories are not restricted to a few pre-selected terms – has not been explored before. Our research has also compared the performance of other approaches such as CLIP-Dissect and GPT-4, revealing that Concept Induction outperforms these methods both quantitatively and qualitatively. However, we acknowledge that there may be some trade-offs between the methods as discussed in Section~\ref{sec:discussion}.
Overall, our line of work aims at comprehensive and conclusive hidden layer analysis for deep learning systems, so that, after analysis, it is possible to
interpret the activations as implicit features of the input that the network has detected, thus opening up avenues to really explain the system’s input-output behavior.

\paragraph{Acknowledgement} The authors acknowledge partial funding under the National Science Foundation grants 2119753 "RII Track-2 FEC: BioWRAP (Bioplastics With Regenerative Agricultural Properties): Spray-on bioplastics with growth synchronous decomposition and water, nutrient, and agrochemical management" and 2333782 "Proto-OKN Theme 1: Safe Agricultural Products and Water Graph (SAWGraph): An OKN to Monitor and Trace PFAS and Other Contaminants in the Nation's Food and Water Systems."

\newpage
\appendix

\bibliographystyle{splncs04}
\bibliography{reference}

\end{document}